\documentclass[11pt]{article}

\usepackage[final]{acl}

\usepackage{times}
\usepackage{latexsym}

\usepackage{amsmath}
\usepackage{amssymb}

\usepackage[T1]{fontenc}

\usepackage[utf8]{inputenc}

\usepackage{microtype}

\usepackage{inconsolata}

\usepackage{graphicx}
\usepackage{xcolor}
\usepackage{booktabs}

\usepackage{framed}

\usepackage{pifont}              
\usepackage[table,xcdraw]{xcolor} 
\usepackage{graphicx}            

\usepackage{booktabs}
\usepackage{multirow}
\usepackage{makecell}
\usepackage{chngcntr}

\usepackage{booktabs}
\usepackage{tabularx}
\usepackage{makecell}
\usepackage{placeins}

\usepackage{tcolorbox}
\usepackage{setspace}
\usepackage{multicol}
\usepackage{mdframed}

\usepackage{float}

\usepackage{float}

\usepackage{xurl}
\usepackage{hyperref}

\definecolor{mygreen}{HTML}{009900}
\definecolor{myred}{HTML}{CC0000}

\newcommand{\cmark}{\textcolor{mygreen}{\scalebox{0.8}{\ding{51}}}}
\newcommand{\xmark}{\textcolor{myred}{\scalebox{0.8}{\ding{55}}}}

\usepackage{xcolor}
\usepackage{tcolorbox}

\definecolor{csTeal}{HTML}{4A9487}
\definecolor{csCream}{HTML}{FBF7F3}

\newtcolorbox{casepromptbox}[1]{%
colback=csCream,
colframe=csTeal,
boxrule=0.6pt,
arc=2mm,
left=3mm,
right=3mm,
top=2mm,
bottom=2mm,
colbacktitle=csTeal,
coltitle=white,
fonttitle=\bfseries\large,
title={#1}
}

\title{AutoPKG: An Automated Framework for Dynamic E-commerce Product-Attribute Knowledge Graph Construction}

\author{
\textbf{Pollawat Hongwimol}$^1$, \textbf{Haoning Shang}$^1$, \textbf{Chutong Wang}$^1$, \\ 
\textbf{Zhichao Wan}$^1$, 
\textbf{Yi Gao}$^1$, \textbf{Yuanming Li}$^{1}$, \textbf{Lin Gui}$^{1}$, \textbf{Wenhao Sun}$^2$, \textbf{Cheng Yu}$^1$ \\
$^1$Lazada, Alibaba International Digital Commerce Group \\
$^2$Nanyang Technological University, Singapore \\
\texttt{\{pollwat.h, shanghaoning.shn, wangchutong.wct, wanzhichao.wzc,} \\ \texttt{xiheng.gy\}@alibaba-inc.com, }
\texttt{yli056@e.ntu.edu.sg, }
\texttt{lin.gui@u.nus.edu, }\\
\texttt{wenhao006@ntu.edu.sg, }
\texttt{yucheng.yc@lazada.com}
}

\begin{document}
\maketitle

\begin{abstract}
Product attribute extraction in e-commerce is bottlenecked by ontologies that are inconsistent, incomplete, and costly to maintain. We present \textbf{AutoPKG}, a multi-agent Large Language Model (LLM) framework that automatically constructs a Product-attribute Knowledge Graph (PKG) from multimodal product content. AutoPKG induces product types and type-specific attribute keys on demand, extracts attribute values from text and images, and consolidates updates through a centralized decision agent that maintains a globally consistent canonical graph. We also propose an \textbf{evaluation protocol} for dynamic PKGs that measures type/key validity and consolidation quality, as well as edge-level accuracy for value assertions after canonicalization. On a large real-world marketplace catalog \textbf{dataset from Lazada (Alibaba)}, AutoPKG achieves up to 0.953 Weighted Knowledge Efficiency (WKE) for product types, 0.724 WKE for attribute keys, and 0.531 edge-level $F_1$ for multimodal value extraction. Across three public benchmarks, we improve edge-level exact-match \(F_1\) by 0.152 and yield a 0.208 precision gain on the attribute extraction application. Online A/B tests show that AutoPKG-derived attributes increase Gross Merchandise Value (GMV) in Badge (+3.81\%), Search (+5.32\%), and Recommendation (+7.89\%), supporting AutoPKG’s practical value in production.
\end{abstract}

\section{Introduction}

Product attributes serve as critical e-commerce infrastructure by powering faceted navigation, improving search relevance, and supporting recommendation and facilitating semantic product understanding at scale \citep{subramaniam2025ai,10.1145/3534678.3539164,10.1145/3292500.3330989}. However, industrial attribute pipelines remain bottlenecked by the \emph{schema}, as product taxonomies and type-specific attribute keys are often inconsistent across markets and incomplete for long-tail inventory \citep{xu-etal-2019-scaling,zhu-etal-2020-multimodal}. Furthermore, it is costly to maintain under continuous distribution shift and multilingual seller noise \citep{xu-etal-2019-scaling,Dong2020}. Consequently, even strong Product Attribute Value Extraction (PAVE) models frequently operate under outdated or overly narrow attribute lists, which limits coverage and necessitates repeated human ontology work \citep{10.1145/3219819.3219839,putthividhya-hu-2011-bootstrapped}.

\begin{table*}[t]
\centering
\small
\setlength{\tabcolsep}{3.8pt} 
\scalebox{0.9}{
\begin{tabular}{lllcccccccc}
\toprule
\textbf{Name} & \textbf{Type} & \textbf{Available} & \textbf{Domain} &
\textbf{\#Nodes} & \textbf{\#Edges} & \textbf{\#Types} & \textbf{\#Attrs} &
\textbf{AT} & \textbf{AK} & \textbf{MM} \\
\midrule

\multicolumn{11}{l}{\textit{\textbf{Frameworks / Systems} (construct \& maintain a KG)}}\\
\midrule
AutoKnow \citep{Dong2020}   & Framework & Closed  & E-commerce  & $>$1B   & $>$1B  & 11K  & >1K  & \cmark & \cmark & \xmark \\
AliCG \citep{10.1145/3447548.3467057}     & Framework & Closed  & E-commerce  & $>$5M & 13.5M  & --  & --   & \cmark & \cmark & \xmark \\
FolkScope \citep{yu-etal-2023-folkscope}  & Framework & Open    & E-commerce  & 1.3M    & 12.8M & --   & --  & \xmark & \cmark & \xmark \\
COSMO \citep{Yu2024}     & Framework & Closed  & E-commerce  & 6.3M    & 29M   & 18   & --  & \cmark & \cmark & \xmark \\
AutoSchemaKG \citep{bai2025autoschemakgautonomousknowledgegraph} & Framework & Open    & General & 900M    & 5.9B  & --   & --  & \cmark & \cmark & \xmark \\

\midrule
\multicolumn{11}{l}{\textit{\textbf{KG Resources} (released graphs)}}\\
\midrule
AliCoCo \citep{10.1145/3318464.3386132}    & KG Resource & Closed & E-commerce  & $>$3B & $>$400B & -- & 20 & \xmark & \xmark & \xmark \\
AliCoCo2 \citep{10.1145/3447548.3467203}   & KG Resource & Closed & E-commerce  & $>$160M & $>$800M & --  & 20  & \xmark & \xmark & \xmark \\
MMpedia \citep{10.1007/978-3-031-47243-5_2}    & KG Resource & Open   & General & 2.6M & 19.5M & --   & --  & \xmark & \xmark & \cmark \\
MedKGent \citep{zhang2025medkgentlargelanguagemodel}  & KG Resource & Open   & Medical & 156K & 3.0M  & --   & --   & \xmark & \cmark & \xmark \\

\midrule
\multicolumn{11}{l}{\textit{\textbf{Datasets / Benchmarks} (PAVE)}}\\
\midrule
OpenTag \citep{10.1145/3219819.3219839}     & Dataset & Partial & E-commerce  & -- & 13K   & 3    & 5    & \xmark & \cmark & \xmark \\
AE-110K \citep{xu-etal-2019-scaling}     & Dataset & Open    & E-commerce  & -- & 110K & 1  & 4 & \xmark & \xmark & \xmark \\
MEPAVE \citep{zhu-etal-2020-multimodal}       & Dataset & Open    & E-commerce & 121K  & 87K & 7  & 26   & \xmark & \xmark & \cmark \\
MAVE \citep{10.1145/3488560.3498377}       & Dataset & Open    & E-commerce  & $>$2M & 3M & 1.3K & 2.5K  & \xmark & \xmark & \xmark \\
ImplicitAVE \citep{zou-etal-2024-implicitave}& Dataset & Open    & E-commerce  & 70K  & --   & 5   & 25   & \xmark & \xmark & \cmark \\

\midrule
\textbf{AutoPKG (Ours)} & \textbf{Framework} & \textbf{Open} & \textbf{E-commerce} &
\textbf{130K} & \textbf{560K} & \textbf{17K} & \textbf{16K} &
\cmark & \cmark & \cmark \\
\bottomrule
\end{tabular}
}
\caption{
Comparison of relevant frameworks, resources and datasets. 
\textbf{\#Types} and \textbf{\#Attrs} denote the number of product categories and unique attribute keys, respectively. 
\textbf{AT}: automatic type induction; \textbf{AK}: automatic key discovery; \textbf{MM}: multimodal value extraction.
}
\label{tab:pkg_comparison_final}
\end{table*}

Table~\ref{tab:pkg_comparison_final} highlights a gap in existing work. While prior e-commerce Knowledge Graph (KG) \emph{frameworks} such as AutoKnow, AliCG, and COSMO demonstrate large-scale construction pipelines, they typically assume or rely on a governed schema while focusing less on \emph{open-ended} type/key induction, multimodal evidence, and continual update. Similarly, open datasets and benchmarks including OpenTag, AE-110K, MEPAVE, MAVE, and ImplicitAVE have advanced PAVE models. However, these resources do not address dynamic schema evolution and canonical KG maintenance as a first-class problem. Furthermore, general-purpose LLM-based schema/KG induction frameworks such as AutoSchemaKG are not tailored to the specific constraints of e-commerce product attributes. Crucially, no prior \emph{open} framework jointly supports \textbf{automatic type induction (AT)}, \textbf{automatic key discovery (AK)}, and \textbf{multimodal value extraction (MM)} within a continually updated PKG.

We present \textbf{AutoPKG}, an automated multi-agent LLM framework \citep{guo2024largelanguagemodelbased} that constructs and continually evolves a PKG from multimodal marketplace listings. Starting from an empty KG, AutoPKG induces \textsc{ProductType} nodes on demand and proposes type-specific \textsc{AttributeKey} nodes. It then extracts attribute values from text and multiple images, finally consolidating all updates into a PKG. A key design element is the centralized \emph{Knowledge Graph Decision agent} (abbreviated as KGD throughout this paper), which acts as the \emph{sole write interface} to the KG. Upstream agents propose edits, whereas KGD resolves them through a constrained action space (\texttt{ADD}/\texttt{MERGE}/\texttt{REPLACE}/\texttt{DISCARD}) using retrieved PKG context. This approach is similar in spirit to validator-based KG construction but operationalized as explicit edit decisions to ensure continual canonicalization \citep{boylan2024kgvalidatorframeworkautomaticvalidation,bian2025llmempoweredknowledgegraphconstruction}.

We also introduce an evaluation protocol for \emph{dynamic PKGs} that measures validity and consolidation quality for induced product types and attribute keys, as well as edge-level precision/recall/F1 for value assertions after canonicalization. On our large real-world catalog dataset, AutoPKG achieves up to 0.953 WKE for product types, 0.724 WKE for attribute keys, and 0.531 edge-level F1 for multimodal value extraction. 
Online A/B tests further demonstrate production impact: AutoPKG-derived attributes improve GMV in Badge (+3.81\%), Search (+5.32\%), and Recommendation (+7.89\%), while delivering no statistically significant change in Filter (+0.26\%).
Furthermore, across three public benchmarks, AutoPKG improves edge level exact-match \(F_1\) by \(+0.152\) and yields a \(+0.208\) (weighted average) precision gain in the downstream PAVE application.

The contributions of this paper are as follows:
\begin{itemize}
\item We propose \textbf{AutoPKG}, the first automated multi-agent framework that unifies automatic type induction, key discovery, and multimodal value extraction within an incrementally evolving PKG, eliminating the need for a fixed, manually maintained taxonomy.
\item We introduce a comprehensive \textbf{evaluation protocol for dynamic PKGs} that goes beyond standard knowledge extraction metrics, specifically measuring schema validity and consolidation quality alongside traditional edge-level correctness.
\item We release a \textbf{large-scale multimodal dataset} comprising 37K Lazada online products and its corresponding KG with 130K nodes and 560K edges across 17K product types. This is the most diverse open-source dataset for e-commerce PAVE that supports multi-image reasoning and automatic schema induction.
\end{itemize}

\begin{figure*}[t]
\centering
\includegraphics[width=\linewidth]{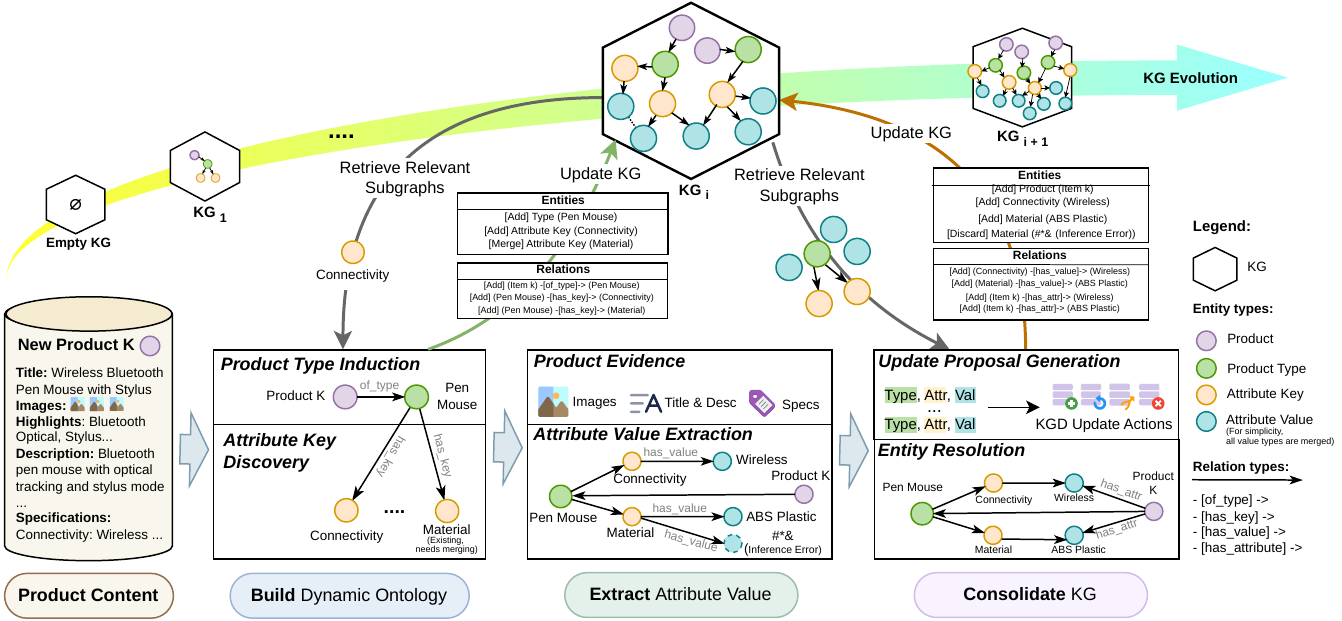}
\caption{AutoPKG overview. From an initially empty KG, the system (1) induces product types and type-specific attribute keys from the listing text, (2) extracts attribute values from the text and images, and (3) consolidates all proposed edits through KGD to maintain a single canonical KG.}
\label{fig:AutoPKG}
\end{figure*}

\section{Related Work}

\paragraph{KG/PKG Construction and Schema Evolution.}
Industrial e-commerce KGs/PKGs are typically built via large-scale extraction plus canonicalization and governance, often emphasizing pipeline scalability rather than open-ended schema evolution \citep{Dong2020,10.1145/3447548.3467057,Yu2024}. 
In parallel, research benchmarks have accelerated extraction modeling but usually assume a predefined label space and do not evaluate continual schema growth \citep{10.1145/3219819.3219839,zou-etal-2024-implicitave}. 
Work on LLM-based KG construction has recently explored schema induction and iterative refinement, but is often batch-oriented and predominantly text-centric 
\citep{boylan2024kgvalidatorframeworkautomaticvalidation}. 
AutoPKG differs by tightly coupling on-demand type/key induction with multimodal value extraction and by treating canonicalization decisions as explicit constrained edits applied continually.

\paragraph{PAVE.}
Earlier approaches typically assume a curated taxonomy and train supervised taggers or classifiers, which are expensive to maintain and brittle under long-tail categories and evolving seller terminology \citep{putthividhya-hu-2011-bootstrapped,xu-etal-2019-scaling}. 
Recent work reframes PAVE as instruction-following or generation with LLMs using schema-constrained prompting, category-aware prompts, and distillation/fine-tuning to improve robustness on noisy product text \citep{brinkmann2024extractgptexploringpotentiallarge,gong2024multilabelzeroshotproductattributevalue,hongwimol-etal-2025-gavel}.
Graph and hypergraph-based approaches have also emerged, leveraging Graph Neural Networks to capture complex structural dependencies in text \citep{potta-etal-2024-attrisage}, or employing heterogeneous hypergraphs to model interrelations between text and image nodes \citep{10.1145/3696410.3714714}.
For visually grounded or implicit attributes (e.g., color, style, pattern), multimodal extraction leverages product images to complement text-only signals \citep{10.1145/3488560.3498377,zou-etal-2024-implicitave,zou-etal-2024-eiven}. 
Retrieval-augmented extraction further improves precision by grounding predictions in similar items or value inventories \citep{zou2025multivalueproductretrievalaugmentedgenerationindustrial}. 
However, most PAVE pipelines still treat the type--key schema as fixed (or manually governed); AutoPKG instead makes schema induction and consolidation first-class and continuous.


\section{AutoPKG Framework}

AutoPKG maintains a PKG ($\mathcal{G}$) and comprises four agents: type-induction agent, key-discovery agent, attribute value extraction agent, and KGD. Figure~\ref{fig:AutoPKG} illustrates the evolution process of $\mathcal{G}$. For each incoming product listing (title, description, specifications, and images), specialized agents propose (i) a product type, (ii) a type-specific attribute-key set, and (iii) item-level attribute values. All updates are mediated by KGD, which performs constrained PKG edits to ensure global consistency. See the Multi-Agent workflow in Appendix~\ref{app:Multi-Agent Workflow}. All prompts used in this section are provided in Appendix~\ref{app:prompts}.

\subsection{PKG Schema}
\label{sec:pkg-schema}
The graph $\mathcal{G}$, maintained by AutoPKG, employs a minimal schema designed for continual expansion under noisy marketplace data (see Appendix~\ref{sec:appendix_pkg_example} for an example subgraph). $\mathcal{G}$ contains three core node types: \textsc{Product} nodes represent marketplace listings, \textsc{ProductType} nodes represent canonical product types induced from listing content and consolidated by KGD, \textsc{AttributeKey} nodes represent canonical attribute keys that define the type-specific schema. Moreover, attribute values are represented as \emph{key-typed value nodes}, each \textsc{Value} node is typed by exactly one \textsc{AttributeKey} and is not treated as an independent ontology type.

AutoPKG uses four edge types that separate schema structure from item facts. Two schema edges represent applicable attribute keys for each product type and valid attribute values for each key: \textsc{(ProductType, has\_key, AttributeKey)}, and \textsc{(AttributeKey, has\_value, Value)}. Two instance edges record product-level facts: \textsc{(Product, of\_type, ProductType)} links an item to its inferred canonical type, while \textsc{(Product, has\_attribute, Value)} asserts an extracted attribute value for that item. This design enforces type checking (values must be licensed by the product type via its keys) and enables canonical value reuse across items after KGD consolidation.

\subsection{Dynamic Ontology Population (Types and Keys)}
\label{sec:ontology}

AutoPKG does not require a pre-defined taxonomy; instead, it populates the ontology in a dynamic manner. For each product, the type-induction agent proposes a canonical \textsc{ProductType} with a brief description from listing text. Conditioned on the inferred type, the key-discovery agent proposes a type-specific \textsc{AttributeKey} table with short definitions and representative value examples. All proposed types/keys are consolidated via KGD.

\subsection{Multimodal Attribute Value Extraction}
\label{sec:pave}

For a product $p$, KGD retrieves the most similar \textsc{ProductType} and corresponding \textsc{AttributeKey} table from $\mathcal{G}$. Then for each attribute key $k$ in the table, the attribute value agent proposes value $v$ assertions using title, description, specifications, and images, producing edges of the form \textsc{($p$, has\_attribute, $v$)} with implicit key type $\mathrm{type}(v)=k$. KGD then canonicalizes and filters these proposals and writes approved edits to $\mathcal{G}$.

\subsection{KGD: The Sole Write Interface}
\label{sec:kgd-agent}

AutoPKG centralizes all KG updates in a single KGD. Upstream agents may propose new product types, attribute keys, values, or item-level edges, but\ only KGD is allowed to write. For each proposal, the agent is given (i) the candidate content and (ii) retrieved local KG neighborhood nodes, and it chooses one constrained edit action: \texttt{ADD} (create a new node/edge), \texttt{MERGE} (put the candidate into an existing canonical node as a synonym/variant), \texttt{REPLACE} (promote the candidate surface form as the new canonical label), or \texttt{DISCARD} (reject invalid, vague, or weakly grounded proposals).

To construct the neighborhood context, we perform dense retrieval over canonical KG node names within the same node type. We embed all canonical names using \texttt{Qwen3-Embedding-0.6B} and retrieve the top-$k$ most similar nodes and provide KGD with their identifiers, names, and available metadata (e.g., descriptions, synonyms). This architecture makes KG growth controllable and consistent by forcing every update through canonicalization-aware, deduplication-focused decisions.

\section{Evaluation Methodology}
\label{sec:eval}

We evaluate AutoPKG as a \emph{dynamic} PKG framework, where model outputs are not only extracted, but also \emph{written into and consolidated within} an incremental evolving PKG. Therefore, our evaluation captures:
(i) semantic validity of induced schema (types/keys),
(ii) consolidation quality under canonicalization,
and (iii) edge-level correctness of value assertions after normalization. Human verification is used to evaluate type/key validity and edge correctness. Sampling sizes and annotation guidelines are provided in Appendix~\ref{app:human-annotation}.

\subsection{Dataset}
\label{sec:dataset}

We sample 323K product listings from a large Southeast Asian e-commerce marketplace by selecting up to 20 items per leaf category per country to maximize type and attribute diversity. Each listing contains (i) a product title, (ii) seller-provided description and highlights, (iii) structured specifications, and (iv) up to ten images. This setting reflects real marketplace noise that challenges continual schema induction and extraction.
We will release a vetted subset and a corresponding PKG snapshot containing 130K nodes and 560K edges, derived from the 323K sampled listings after filtering and consolidation.\footnote{\url{https://github.com/Product-Understanding-Lazada-Alibaba/AutoPKG.git}} See details in Appendix~\ref{app:data-release}.

\subsection{Models}
\label{sec:models}

We evaluate a diverse suite of instruction-tuned LLMs, categorizing them by modality and scale. For text-only tasks (Sections~\ref{sec:ontology} and \ref{sec:kgd-agent}), we employ LLMs including Qwen3 \citep{qwen3}, Gemma-3 and Gemma-3n \citep{gemma_2025}, Llama-3.2 \citep{llama3herdmodels}, Kimi \citep{team2025kimi}, GroveMoE \citep{GroveMoE}, and SmolLM3 \citep{bakouch2025smollm3}; Attribute value extraction from multimodal product listings is handled by MLLMs, specifically Qwen3-VL \citep{Qwen3-VL}, Gemma-3 \citep{gemma_2025}, and Kimi-VL \citep{kimiteam2025kimivltechnicalreport}. For a controlled comparison, all backbones are evaluated with an identical zero-shot prompt template and standardized hyperparameters. See implementation details in Appendix~\ref{app:hardware}.

\subsection{KGD Decision Evaluation via Multi-LLM Consensus}
\label{sec:kgd-eval}

We evaluate KGD model candidates using a panel of five frontier LLM judges: DeepSeek-V3.1 \citep{deepseekai2025deepseekv3technicalreport}, Gemini-2.5-Pro \citep{Comanici2025Gemini2P}, GPT-5 \citep{openai_gpt5_2025}, Kimi-K2 \citep{kimiteam2025kimik2openagentic}, and Qwen3-Max \citep{qwen3}. 
For each decision instance, judges select an edit action given the candidate and retrieved PKG context; the reference label is the majority vote ($\geq$3/5). Candidate model performance is reported as accuracy relative to this consensus, and we assess the reliability of the judging panel through pairwise Cohen’s $\kappa$ scores. We further validate the reliability of the LLM-judge consensus via human annotation on a subset, observing strong agreement (N = 2{,}000, $\kappa = 0.837$; see Appendix~\ref{app:judge-agreement}, Table~\ref{tab:llm-judge-kappa}).

Due to the absence of a pre-existing dataset for initial PKG states, we synthesize an experimental corpus by generating \textsc{ProductType} candidates using \texttt{Qwen3-Next-80B-A3B-Instruct}. These candidates are subsequently integrated into empty KGs via KGD under varied configurations. Then, we perform random sampling from this synthesized pool to construct a balanced dataset for evaluation.

\subsection{Product Type Metrics}
\label{sec:type-metrics}

Let $N$ be the number of products processed in evaluation, and $V$ be the number of
canonical \textsc{ProductType} nodes after KGD consolidation. We report:
Acceptance ($R_{\text{acc}}$), the fraction of sampled product type predictions
judged valid by humans;
Compression ($R_{\text{comp}} = 1 - V/N$), the redundancy reduction achieved by canonicalization; and
Coverage ($R_{\text{cov}}$), the fraction of products assigned to a valid
canonical type. 
We summarize overall performance using Weighted Knowledge Efficiency ($\text{WKE}_{type}$), a weighted harmonic mean:
\[
\text{WKE}_{type} =
\frac{w_1 + w_2 + w_3}{\frac{w_1}{R_{\text{acc}}} + \frac{w_2}{R_{\text{comp}}} + \frac{w_3}{R_{\text{cov}}}}.
\]
This formulation prevents over-aggressive merging from inflating scores. In practice, the weights $w_1=3, w_2=1, w_3=1$ yield the best balance between semantic validity, compression, and coverage, as detailed in Appendix~\ref{app:wke-weighting}.

\subsection{Attribute Key Metrics (Probabilistic)}
\label{sec:key-metrics}

Exhaustively enumerating ground-truth attribute keys for open-vocabulary, type-specific schemas is infeasible. So we adapt a probabilistic evaluation protocol, inspired by probabilistic precision/recall for incomplete labeling and multi-system pooling \citep{10.1007/978-3-642-36824-0_15,park2023probabilistic}.

Let $\mathcal{K}_{\text{total}}$ be the pooled set of unique canonical keys obtained from the top-performing key-generation models (selected by acceptance) after consolidation via KGD. For each model $M_j$, we estimate a reliability prior $P(M_j)$ using human key acceptance on sampled outputs, i.e., $P(M_j)=R_{\text{acc}}(M_j)$.

For each canonical key $k\in\mathcal{K}_{\text{total}}$, let $\text{predictors}(k)$ be the set of models that proposed $k$. We estimate its probability of belonging to the latent ground truth under a Noisy-OR assumption:
\[
P(k\in \text{GT}) = 1-\prod_{M_j\in\text{predictors}(k)} (1-P(M_j)).
\]
We compute probabilistic true positives for $M_j$ as
$\text{TP}(M_j)=\sum_{k\in\mathcal{K}_j}P(k\in\text{GT})$, where $\mathcal{K}_j$ is
the canonical key set produced by $M_j$. Let $|\widehat{\text{GT}}|=
\sum_{k\in\mathcal{K}_{\text{total}}}P(k\in\text{GT})$. Then:
\[
\text{P-Prec}(M_j)=\frac{\text{TP}(M_j)}{|\mathcal{K}_j|},\quad
\text{P-Rec}(M_j)=\frac{\text{TP}(M_j)}{|\widehat{\text{GT}}|}.
\]
Similar to Section~\ref{sec:type-metrics}, We report WKE with weights $w_1=3, w_2=1, w_3=1$:
\[
\text{WKE}_{key} =
\frac{w_1 + w_2 + w_3}{\frac{w_1}{R_{\text{acc}}} + \frac{w_2}{\text{P-Prec}} + \frac{w_3}{\text{P-Rec}}},
\]
where the weighting rationale is the same as that of $\text{WKE}_{type}$ (Appendix~\ref{app:wke-weighting}).

\subsection{Attribute Value Metrics (Edge-Level)}
\label{sec:value-metrics}

We evaluate attribute value extraction at \emph{KG edge level} after KGD canonicalization. Each predicted assertion has the form \textsc{(Product, has\_attribute, Value)}. Because each \textsc{Value} node is typed by exactly one \textsc{AttributeKey} via \textsc{(AttributeKey, has\_value, Value)} (Section~\ref{sec:pkg-schema}), an assertion is counted as correct only if (i) the value is supported by the listing evidence and (ii) the value is attached to the intended key type after canonicalization.

\paragraph{Single Canonical Evaluation Graph}
To avoid penalizing surface-form variation (e.g., ``Grey'' vs.\ ``Gray'') and to score all models in the same canonical space, we build an evaluation KG by taking the union of value-edge proposals from all evaluated extractors and letting KGD mediate all writes. Each model is then evaluated on the subset of canonicalized edges originating from that model. Human annotators label instantiated edges as Correct/Incorrect using full listing evidence. We report precision, recall, and F1 based on these labeled edges, with true/false positives/negatives defined over canonicalized \textsc{(Product,has\_attribute,Value)} edges.

\subsection{Public Dataset Evaluation: Edge-Level PKG Assessment}
\label{sec:public-eval-kg}

To assess AutoPKG beyond our primary benchmark, we conduct an edge-level evaluation on three public datasets: ImplicitAVE, MAVE, and AE-650k. Using the same canonicalized evaluation procedure described in Section~\ref{sec:value-metrics}, we construct a single evaluation KG per dataset by taking the value assertions from multiple high-performing LLMs from Section \ref{sec:results-value} and letting KGD determine the final node/edge instantiations. Subsequently, the flagship Qwen3-VL-235B-A22B-Instruct model is used as an AI annotator to generate high-quality reference graphs as silver ground truth. We validate the reliability of these annotations via human evaluation on a sampled subset, observing strong agreement (N = 1{,}000, Cohen’s Kappa = 0.8297; see Appendix~\ref{app:judge-agreement-open}). Finally, we compare subgraphs from public datasets and ours in the shared canonical space and report macro-averaged precision, recall, and F1 (details: Appendix~\ref{app:Open-Dataset-Experimental}, Table~\ref{tab:kg_finegrained_node_quality}).

\subsection{Public Dataset Evaluation: PKG-Augmented PAVE}
\label{sec:public-eval-pave}
We additionally evaluate AutoPKG in a PKG-augmented PAVE setting on the superset of three public datasets used in Section~\ref{sec:public-eval-kg} (ImplicitAVE, MAVE, and AE-650k). For each dataset, we use the PKG produced in Section~\ref{sec:public-eval-kg} as a retrieval source for graph retrieval-augmented generation (G-RAG) \citep{hu2025graggraphretrievalaugmentedgeneration} when performing the PAVE task. We then evaluate the PAVE outputs with Qwen3-VL-235B-A22B-Instruct serving also as an AI annotator to construct silver ground truth annotations, and report macro-averaged precision, recall, and F1. Full evaluation and metric computation details are provided in Appendix~\ref{app:AP-PAVE-Task-Assessment}. See human-verified silver GT judgments in Appendix~\ref{app:judge-agreement-open}.

\section{Results}
\label{sec:results}

We report results for (i) KGD under multi-LLM judging, (ii) product-type induction and consolidation, (iii) attribute-key discovery under probabilistic evaluation, (iv) multimodal attribute value extraction at the item--value edge level, (v) edge-level evaluation on public datasets, (vi) PKG-augmented PAVE task in E-commerce, and (vii) an industrial online A/B deployment. Latency is reported as wall-clock elapsed time multiplied by the number of GPUs used, formatted as \texttt{HH:MM:SS}, under consistent hardware settings (Appendix~\ref{app:hardware}).

\subsection{KGD: Decision Fidelity and Judge Agreement}
\label{sec:results-kgd}

Table~\ref{tab:judge-accuracy} evaluates candidate backbones for KGD using a 5-judge panel and a 3/5 majority-vote consensus label. Qwen3-Next-80B-A3B achieves the highest consensus accuracy (0.764), followed by Qwen3-30B-A3B (0.734). The drop for smaller models is sharp (e.g., Llama-3.2-3B at 0.384 and SmolLM3-3B at 0.123), suggesting that constrained KG edits still require strong semantic discrimination: mistakes in \texttt{MERGE} vs.\ \texttt{ADD} (or overly confident \texttt{REPLACE}) have persistent downstream cost because they shape the canonical ontology used for retrieval and extraction.

\begin{table}[t]
\centering
\small
\setlength{\tabcolsep}{5pt}
\begin{tabular}{l c c}
\toprule
\textbf{Model} & \textbf{Latency} ($\downarrow$) & \textbf{Acc.} ($\uparrow$) \\
\midrule
Qwen3-Next-80B-A3B-Instruct       & 4:37    & \textbf{0.764} \\
Qwen3-30B-A3B-Instruct-2507       & 1:51    & 0.734 \\
Qwen3-4B-Instruct-2507            & 1:38    & 0.666 \\
Kimi-Linear-48B-A3B-Instruct      & 37:28   & 0.610 \\
GroveMoE-Inst                     & 4:39:28 & 0.594 \\
Llama-3.2-3B-Instruct             & 2:14    & 0.384 \\
Gemma-3n-E4B-it                   & 3:06    & 0.233 \\
SmolLM3-3B                        & 3:23    & 0.123 \\
\bottomrule
\end{tabular}
\caption{KGD backbone accuracy against a 5-LLM majority-vote reference (3/5). Time is wall-clock elapsed time; higher accuracy and lower latency indicate better performance.}
\label{tab:judge-accuracy}
\end{table}

\begin{table*}[t]
\centering
\small
\setlength{\tabcolsep}{5pt}

\begin{tabular}{l r r c c c c}
\toprule

\multicolumn{7}{l}{\textbf{Product-type Induction}} \\
\midrule
\textbf{Model} & \textbf{Unique Types} & \textbf{Node Size} &
\textbf{Accept.} ($\uparrow$) & \textbf{Comp.} ($\uparrow$) & \textbf{Cov.} ($\uparrow$) &
\textbf{WKE} ($\uparrow$) \\
\midrule
Qwen3-4B-Instruct-2507        & 80{,}806  & 16{,}692 & 0.952 & 0.948 & 0.960 & \textbf{0.953} \\
Qwen3-30B-A3B-Instruct-2507   & 73{,}620  & 20{,}711 & \textbf{0.954} & 0.936 & 0.960 & 0.952 \\
Qwen3-Next-80B-A3B-Instruct   & 81{,}565  & 21{,}608 & \textbf{0.954} & 0.933 & 0.951 & 0.949 \\
GroveMoE-Inst                 & 51{,}957  & 14{,}787 & 0.948 & 0.954 & 0.952 & 0.950 \\
Kimi-Linear-48B-A3B-Instruct  & 101{,}062 & 25{,}311 & 0.938 & 0.946 & 0.922 & 0.940 \\
Gemma-3n-E4B-it               & 37{,}645  & 10{,}603 & 0.936 & \textbf{0.967} & \textbf{0.986} & 0.952 \\
Llama-3.2-3B-Instruct         & 55{,}317  & 17{,}470 & 0.628 & 0.946 & 0.945 & 0.726 \\
SmolLM3-3B                    & 198{,}814 & 21{,}649 & 0.646 & 0.933 & 0.613 & 0.682 \\
\midrule

\multicolumn{7}{l}{\textbf{Attribute-key Discovery}} \\
\midrule
\textbf{Model} & \textbf{Unique Keys} & \textbf{Node Size} &
\textbf{Accept.} ($\uparrow$) & \textbf{P-Prec.} ($\uparrow$) & \textbf{P-Rec.} ($\uparrow$) &
\textbf{WKE} ($\uparrow$) \\
\midrule
Qwen3-235B-A22B-Instruct-2507  & 23{,}005 & 7{,}994  & \textbf{0.956} & 0.990 & 0.363 & \textbf{0.724} \\
Qwen3-30B-A3B-Instruct-2507    & 34{,}738 & 10{,}634 & 0.794 & 0.934 & \textbf{0.474} & 0.718 \\
Kimi-Linear-48B-A3B-Instruct   & 24{,}002 & 8{,}560  & 0.806 & 0.946 & 0.367 & 0.667 \\
Qwen3-4B-Instruct-2507         & 23{,}689 & 7{,}007  & 0.822 & 0.950 & 0.353 & 0.664 \\
Qwen3-Next-80B-A3B-Instruct    & 17{,}432 & 6{,}663  & 0.942 & \textbf{0.991} & 0.273 & 0.636 \\
Gemma-3n-E4B-it                & 32{,}504 & 7{,}924  & 0.628 & --    & --    & --    \\
Llama-3.2-3B-Instruct          & 29{,}059 & 4{,}239  & 0.534 & --    & --    & --    \\
SmolLM3-3B                     & 10{,}211 & 1{,}905  & 0.494 & --    & --    & --    \\
\bottomrule
\end{tabular}

\caption{KGD-based consolidation quality for two induction tasks. 
Top block: \textit{Product-type induction} where \textit{Unique Types} are raw predicted strings and \textit{Node Size} is the number of canonical type nodes after consolidation; \textit{Accept.} is the human validity rate of per-product type predictions (pre-consolidation); \textit{Comp.} and \textit{Cov.} are computed after consolidation; \textit{WKE} aggregates these metrics.
Bottom block: \textit{Attribute-key discovery} where \textit{Unique Keys} are raw generated keys and \textit{Node Size} is the number of canonical key nodes after consolidation; \textit{Accept.} is the human validity rate on sampled per-product generated keys (pre-consolidation); \textit{P-Prec.}, \textit{P-Rec.}, and \textit{WKE} are computed using the probabilistic protocol with a pooled key set from the top models.}
\label{tab:type-and-attrkey-quality}
\end{table*}





\subsection{Product Type Induction: Validity under Aggressive Consolidation}
\label{sec:results-type}

Table~\ref{tab:type-and-attrkey-quality} (top block) reports type induction quality after KGD-mediated consolidation. Strong models produce highly valid canonical types, with acceptance around 0.94--0.95 for most candidates. Notably, the system collapses a very large space of raw type strings into a much smaller set of canonical nodes, yielding high compression (0.933--0.967) while maintaining high coverage (0.945--0.986). This combination indicates that consolidation is not merely deleting long-tail types; rather, it is mostly removing surface-form fragmentation (pluralization, spelling variants, seller-specific phrasing) while still assigning almost all products to a valid type.

WKE summarizes this trade-off. Qwen3-4B achieves the highest WKE (0.953), narrowly ahead of Qwen3-30B (0.952). Interestingly, Gemma-3n achieves the strongest compression and coverage but slightly lower acceptance, consistent with a more aggressive merge tendency: when merges become too permissive, acceptance is penalized strongly by WKE. Overall, these results suggest that for type induction, moderate-size backbones can be sufficient if the KGD policy is stable, and that the dominant failure mode is not missing coverage but subtle semantic drift from over-merging.

\subsection{Attribute Key Discovery: High Precision, Recall Limited by Long Tail}
\label{sec:results-key}

Table~\ref{tab:type-and-attrkey-quality} (bottom block) evaluates induced attribute keys using the probabilistic precision/recall protocol in Section~\ref{sec:key-metrics}. Among the evaluated top models, probabilistic precision is consistently high (roughly 0.93--0.99), indicating that once KGD canonicalizes proposals, most surviving keys are meaningful and type-relevant. In contrast, probabilistic recall remains much lower (0.273--0.474), reflecting the open-vocabulary nature of e-commerce attributes: Many useful keys are rare, category-specific, or expressed idiosyncratically.

Qwen3-235B achieves the best WKE (0.724), driven by the strongest acceptance (0.956) and near-ceiling probabilistic precision, making it suitable when ontology growth must be conservative and stable. Qwen3-30B yields the best recall (0.474) and a comparable WKE (0.718), suggesting a complementary role as a higher-coverage proposer if guarded by a strong KGD. Taken together, the key results reinforce that schema evolution is bottlenecked less by filtering invalid keys than by capturing the long tail without inflating the ontology with weakly grounded concepts.

\subsection{Attribute Value Extraction: Precision--Recall Trade-offs}
\label{sec:results-value}

Table~\ref{tab:attr-value-extraction} reports edge-level precision/recall/F1 for multimodal value extraction after KGD canonicalization (Section~\ref{sec:value-metrics}). Overall performance is moderate, reflecting noisy marketplace evidence (incomplete specifications, ambiguous seller texts, and non-diagnostic images). Models exhibit a clear precision--recall trade-off: Qwen3-VL-32B achieves the best F1 (0.531) via the highest recall (0.591), while Gemma-3-27B is most conservative with the highest precision (0.688) but lower recall (0.420), yielding a comparable F1 (0.521).

Latency varies substantially, making model choice a throughput--quality decision. Smaller Qwen3-VL models (4B/8B) provide competitive F1 (0.448--0.482) at much lower runtime. Scaling within a family is not monotonic: the MoE Qwen3-VL-30B-A3B underperforms the dense Qwen3-VL-32B (0.446 vs.\ 0.531), suggesting structured extraction is sensitive to calibration and routing. 

\begin{table}[t]
\centering
\small
\setlength{\tabcolsep}{5pt}
\scalebox{0.93}{
\begin{tabular}{l c r c c c}
\toprule
\textbf{Model} & \textbf{Latency} ($\downarrow$) & \textbf{Prec.} & \textbf{Rec.} & \textbf{F1} ($\uparrow$) \\
\midrule
Qwen3-VL-32B              & 2:59:04 & 0.483 & \textbf{0.591} & \textbf{0.531} \\
Gemma-3-27b-it            & 4:43:12 & \textbf{0.688} & 0.420 & 0.521 \\
Gemma-3-12b-it            & 4:15:01 & 0.626 & 0.396 & 0.485 \\
Qwen3-VL-8B               & 1:50:52 & 0.455 & 0.513 & 0.482 \\
Qwen3-VL-4B               & 1:50:51 & 0.436 & 0.462 & 0.448 \\
Qwen3-VL-30B-A3B          & 2:17:22 & 0.389 & 0.522 & 0.446 \\
Gemma-3n-E4B-it           & 4:56:59 & 0.521 & 0.348 & 0.417 \\
Kimi-VL-A3B               & 5:58:20 & 0.394 & 0.258 & 0.312 \\
\bottomrule
\end{tabular}
}
\caption{Attribute value extraction on the human-labeled set (edge-level) after KGD canonicalization.}
\label{tab:attr-value-extraction}
\end{table}



\subsection{Public Datasets: Edge-Level Results}
The AutoPKG-constructed \(\mathrm{KG}_2\) improves exact-match macro-\(F_1\) over the strongest baseline \(\mathrm{KG}_1\), with Gemma-3-27b-it delivering the largest average gains: \(+0.166\) on \textsc{AttributeKey} nodes and \(+0.231\) on \textsc{AttributeValue} nodes across ImplicitAVE/MAVE/AE--650K (absolute macro-\(F_1\), averaged across datasets), up to \(+0.207\) in precision on the multimodal ImplicitAVE dataset. See details in Table~\ref{tab:kg_finegrained_node_quality}. Overall, the gains align with the pipeline mechanism in AutoPKG.

\subsection{Public Datasets: PKG-Augmented PAVE Results}

Table~\ref{tab:ave-main} presents the PAVE results on three public datasets. PKG-augmented methods outperform the original dataset baselines. Gemma-3-27b-it yielding the best overall gains: improving exact-match macro-$F_{1}$ by \(+0.391\) on \textsc{AttributeKey} and \(+0.445\) on \textsc{Value} on average over the two text-only settings (MAVE and AE-650K), while further increasing ImplicitAVE precision by \(+0.256\) (keys) and \(+0.063\) (values). 
The gains are primarily driven by PKG augmentation: retrieving the relevant KG exposes canonical attribute schemas and candidate values, which constrains generation and improves coverage through graph-supported evidence aggregation.

\begin{table}[t]
\centering
\small
\setlength{\tabcolsep}{4pt}
\resizebox{\columnwidth}{!}{%
\begin{tabular}{@{}l p{2.3cm} ccc ccc@{}}
\toprule
\multirow{2}{*}{Dataset} & \multirow{2}{*}{\makecell[l]{AVE Task\\Comparison}} &
\multicolumn{3}{c}{Attribute Key} & \multicolumn{3}{c}{Attribute Value} \\
\cmidrule(lr){3-5}\cmidrule(lr){6-8}
& & P & R & F1 & P & R & F1 \\
\midrule

\multirow{3}{*}{ImplicitAVE}
& Original Dataset
& 0.714 & -- & -- & 0.890 & -- & -- \\
& Gemma-3-27b-it
& \textbf{0.970} & 0.894 & \textbf{0.924} & \textbf{0.953} & 0.876 & 0.907 \\
& Qwen3-VL-32B
& 0.911 & \textbf{0.951} & 0.924 & 0.897 & \textbf{0.932} & \textbf{0.908} \\
\midrule

\multirow{3}{*}{Mave}
& Original Dataset
& 0.774 & 0.187 & 0.282 & 0.625 & 0.175 & 0.250 \\
& Gemma-3-27b-it
& \textbf{0.855} & 0.663 & 0.726 & \textbf{0.895} & 0.718 & 0.777 \\
& Qwen3-VL-32B
& 0.767 & \textbf{0.751} & \textbf{0.746} & 0.802 & \textbf{0.806} & \textbf{0.792} \\
\midrule

\multirow{3}{*}{AE650k}
& Original Dataset
& 0.805 & 0.291 & 0.407 & 0.774 & 0.238 & 0.344 \\
& Gemma-3-27b-it
& \textbf{0.826} & \textbf{0.597} & \textbf{0.671} & \textbf{0.856} & \textbf{0.631} & \textbf{0.707} \\
& Qwen3-VL-32B
& 0.466 & 0.462 & 0.442 & 0.487 & 0.507 & 0.475 \\
\bottomrule
\end{tabular}%
}
\caption{PAVE performance on three public datasets.}
\label{tab:ave-main}
\end{table}

\subsection{Online Deployment: Filter, Badge, Search, and Recommendation}
\label{sec:results-online}

We deployed AutoPKG-derived attributes in four production applications and conducted A/B tests against the existing system (\textit{Base}) to measure the effect of the AutoPKG-derived attributes (\textit{Test}). Appendix~\ref{app:demos} provides UI demos. In UI-facing settings, AutoPKG attributes are used for \textit{Filter} facets and \textit{Badge} display: 
\textit{Filter} exposes selectable facet values to narrow result sets, while \textit{Badge} provides attribute snippets on the listing card. 
Over 12/01--12/28, \textit{Filter} shows no statistically significant change (GMV +0.26\%), while \textit{Badge} improves GMV by +3.81\%. Badge likely benefits from higher effective coverage since partial attributes can still inform decisions without restricting retrieval.

We also deployed AutoPKG for \textit{Attribute Backfilling} to enrich missing attributes on 110M items. \textit{Base} relies on seller-provided attributes (46\% fill rate), while \textit{Test} adds AutoPKG backfill (70\% fill rate). Over 11/21--12/22, this backfilling improves GMV by +5.32\% in \textit{Search} and +7.89\% in \textit{Recommendation}, suggesting that most value comes from long-tail coverage gains that strengthen downstream retrieval, matching, and ranking.

\subsection{Cost--Quality Trade-off Analysis}
\label{sec:cost-analysis}

\begin{table*}[t]
\centering
\small
\setlength{\tabcolsep}{3.8pt} 
\renewcommand{\arraystretch}{1.1}
\resizebox{\textwidth}{!}{%
\begin{tabular}{l l l l l c c c c c r}
\toprule
\multirow{2}{*}{\textbf{Config}} & \multicolumn{4}{c}{\textbf{Agent Backbones}} & \multicolumn{4}{c}{\textbf{Performance Metrics}} & \multirow{2}{*}{\textbf{Avg}} & \multirow{2}{*}{\textbf{Cost}} \\
\cmidrule(lr){2-5} \cmidrule(lr){6-9}
& \textbf{KGD} & \textbf{Type Ind.} & \textbf{Key Disc.} & \textbf{Value Extr.} & \textbf{Acc$_{\text{KGD}}$} & \textbf{WKE$_{\text{type}}$} & \textbf{WKE$_{\text{key}}$} & \textbf{Edge $F_1$} & \textbf{Qual.}$^\dagger$ & \textbf{Est. (\$/1M)} \\
\midrule
\textit{Minimal} & Qwen3-30B & Qwen3-4B & Qwen3-30B & Qwen3-VL-8B & 0.734 & 0.953 & 0.718 & 0.482 & 0.680 & \$12,012 \\
\textit{Balanced} & Qwen3-30B & Qwen3-4B & Qwen3-30B & Qwen3-VL-32B & 0.734 & 0.953 & 0.718 & 0.531 & 0.703 & \$12,452 \\
\textit{Full} & Qwen3-Next-80B & Qwen3-4B & Qwen3-235B & Qwen3-VL-32B & \textbf{0.764} & \textbf{0.953} & \textbf{0.724} & \textbf{0.531} & \textbf{0.711} & \$14,000 \\
\bottomrule
\end{tabular}%
}
\vspace{-0.5em}
\caption{Detailed cost--quality breakdown across three deployment configurations. 
$^\dagger$\textbf{Avg Qual.}: Harmonic mean of Acc$_{\text{KGD}}$, WKE$_{\text{type}}$, WKE$_{\text{key}}$, and Edge $F_1$. 
Costs are estimated based on batch processing at full GPU utilization on 8$\times$NVIDIA H20 GPUs.}
\label{tab:cost-quality}
\end{table*}

To guide deployment decisions, we evaluate three configuration tiers ranging from latency-optimized to performance-maximized setups. Table~\ref{tab:cost-quality} reports the aggregate quality score (harmonic mean of KGD accuracy, type/key WKE, and edge $F_1$) alongside inference costs estimated on AWS p5e.48xlarge instances (\$4/GPU-hr). See configuration suggestions in Appendix~\ref{app:configuration_s}.

\subsection{Ablation Study: Impact of KGD Context and Multimodality}
\label{sec:ablation}

\begin{table}[t]
\centering
\small
\setlength{\tabcolsep}{4pt}
\begin{tabular}{l c c c}
\toprule
\textbf{Model Variant} & \textbf{WKE$_{\text{type}}$} & \textbf{WKE$_{\text{key}}$} & \textbf{Edge $F_1$} \\
\midrule
\textbf{Full AutoPKG (Ours)} & \textbf{0.952} & \textbf{0.719} & \textbf{0.532} \\
\quad -- w/o retrieval context & 0.886 & 0.680 & 0.490 \\
\quad -- w/o visual evidence & 0.952 & 0.719 & 0.523 \\
\quad -- Both ablations & 0.886 & 0.680 & 0.420 \\
\bottomrule
\end{tabular}
\caption{Ablation study on the impact of KGD retrieval context and multimodal inputs.}
\label{tab:ablation}
\end{table}

To quantify the contribution of key architectural components, we conduct an ablation study removing the retrieved neighborhood context in the KGD agent and multimodal inputs in the value extractor (Table~\ref{tab:ablation}). Removing the KGD retrieval context forces the decision agent to operate without graph consistency checks, leading to a significant drop in schema quality ($\text{WKE}_{type}$ decreases by 0.066 and $\text{WKE}_{key}$ by 0.039) as the agent fails to effectively merge synonyms or detect duplicates, resulting in a fragmented ontology. Disabling multimodal inputs while retaining text-only extraction causes a modest decline in edge-level $F_1$ (from 0.532 to 0.523); while text specifications often contain explicit values, visual evidence remains critical for resolving ambiguities, verifying implicit attributes (e.g., pattern, style), and compensating for noisy seller text. The combined removal of both components yields the lowest performance across all metrics, confirming that AutoPKG's effectiveness relies on the synergy between context-aware canonicalization and multimodal evidence aggregation.

\section{Conclusion}

We introduce AutoPKG, a production-oriented multi-agent framework for building and incrementally evolving product-attribute knowledge graphs from multimodal e-commerce listings. AutoPKG treats product types and attribute keys as dynamic rather than fixed, and centralizes KG updates to maintain global consistency during continual growth. On a large real-world marketplace dataset, we find that type induction can remain highly valid while aggressively consolidating surface variants, that key induction achieves high precision but is recall-limited in the long tail, and that multimodal value extraction exhibits a clear precision--recall trade-off across different backbones. 
Online A/B tests show heterogeneous production impact: \textit{Badge}, \textit{Search}, and \textit{Recommendation} deliver clear GMV gains, while \textit{Filter} shows no statistically significant change, motivating per-vertical gating and precision-oriented guardrails when promoting extracted attributes to ranking- or UI-facing features.
Overall, AutoPKG offers a practical route to reducing manual taxonomy maintenance while keeping a product KG aligned with a fast-evolving catalog.

\section{Limitations}
\label{sec:limitations}


AutoPKG writes value into a incremental  evolving KG, early judgment errors can persist and propagate: an incorrect \texttt{MERGE} may collapse distinct concepts, an overly broad canonical node may distort retrieval context for future products, and an incorrect \texttt{ADD} may inflate the ontology with low-utility or noisy schema elements. Although KGD constrains edits to a limited action space, quality remains sensitive to the retrieved neighborhood and to subtle semantic distinctions (e.g., near-synonyms vs. true synonyms, or homographs across domains).

The schema evaluation is also limited by the open-vocabulary setting. In particular, attribute-key quality cannot be exhaustively labeled; our probabilistic protocol depends on system pooling and model-reliability priors, which may under-represent rare but valuable long-tail keys and may favor keys proposed by multiple similar models. Attribute value extraction is additionally constrained by noisy marketplace evidence, such as inconsistent specifications, missing or low-quality images, and multilingual/code-mixed text, leading to both false positives (spurious assertions) and false negatives (missed implicit attributes).

\section{Ethics and Risks}
\label{sec:ethics-and-risks}
\paragraph{Ethics.} This work uses marketplace catalog content and internal employee annotation for product understanding and quality evaluation. It does not involve collection of user data or intervention with human subjects. The study has been reviewed under internal research governance procedures and deemed exempt from external IRB review under applicable policies.

\paragraph{Potential Risks.} As an automated framework powered by large language models, AutoPKG may produce incorrect, biased, or misleading attributes that harm user experience (e.g., wrong facet filters or badges), disadvantage sellers (e.g., mischaracterized product properties), or degrade ranking and recommendation when used as features. Quality may vary across languages, regions, and long-tail categories, leading to uneven coverage and disparate impact. While our released subset undergoes internal approval, includes PII redaction where applicable, and is subject to content-safety screening for both listings and images, it may still contain sensitive commercial information (e.g., brands or seller-provided identifiers). We therefore recommend conservative deployment: gate high-impact keys with precision-oriented thresholds, monitor drift and error patterns by vertical and language, and periodically audit canonicalization decisions that affect large fractions of traffic.

\bibliography{custom}

\appendix
\counterwithin{table}{section}
\counterwithin{figure}{section}
\renewcommand{\thetable}{\thesection.\arabic{table}}
\renewcommand{\thefigure}{\thesection.\arabic{figure}}

\section{Product Knowledge Graph Example}
\label{sec:appendix_pkg_example}

\begin{figure*}[ht]
\centering
\includegraphics[width=0.9\textwidth]{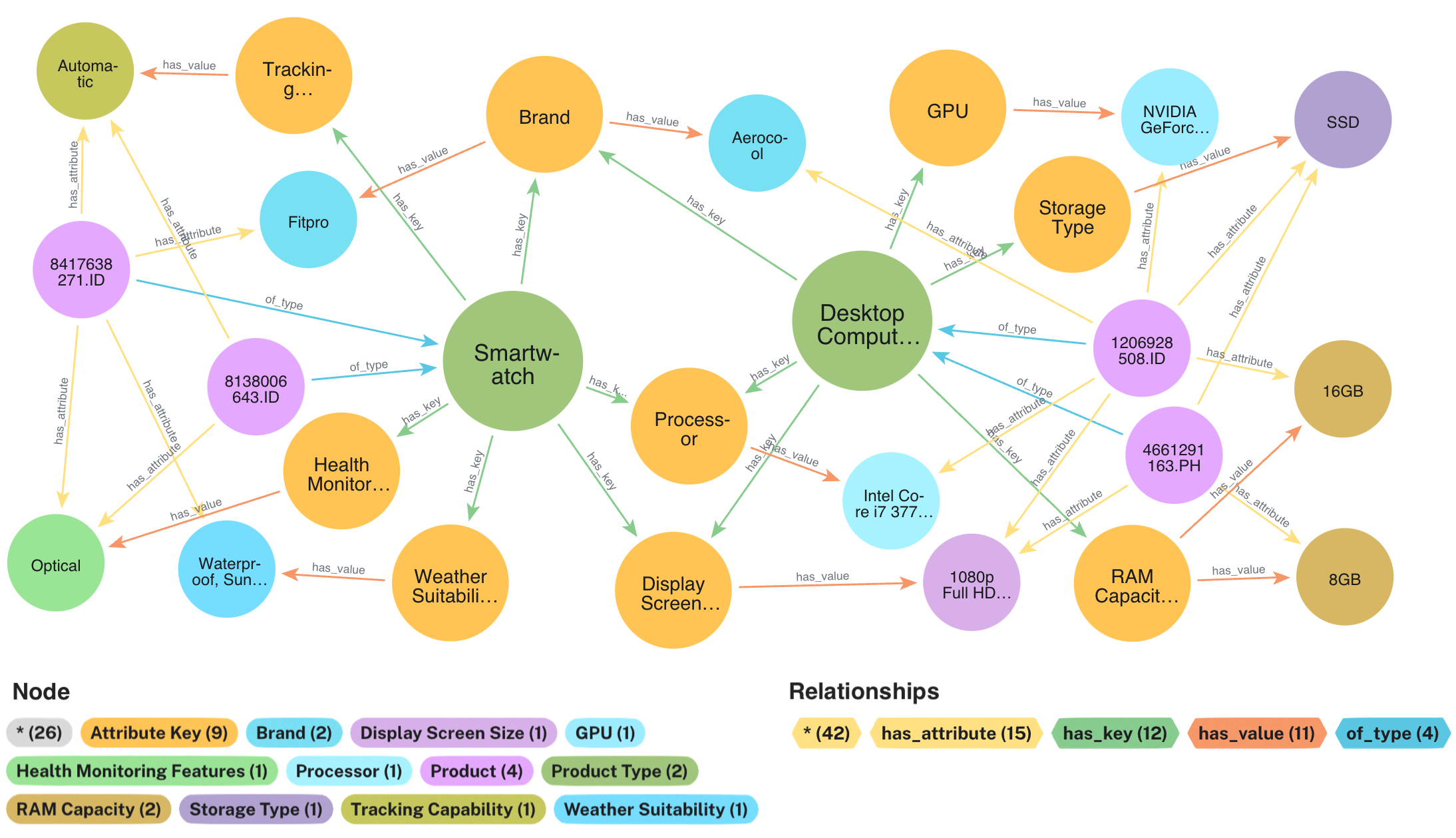}
\caption{Example subgraph of a PKG constructed from extracted product attributes.}
\label{fig:subgraph}
\end{figure*}

This appendix provides a concrete example of the Product-attribute Knowledge Graph (PKG) produced by AutoPKG. Figure~\ref{fig:subgraph} visualizes a local neighborhood after KGD canonicalization.

In the figure, large green nodes denote canonical \textsc{ProductType} concepts (e.g., \emph{Smartwatch}, \emph{Desktop Computer}). Orange nodes denote \textsc{AttributeKey} slots that are valid for a given product type (e.g., \emph{Brand}, \emph{GPU}, \emph{RAM Capacity}, \emph{Storage Type}). Purple nodes denote individual \textsc{Product} items. The remaining nodes correspond to canonical \textsc{Value} entities (e.g., \emph{SSD}, \emph{16GB}, \emph{NVIDIA GeForce}), which are reused across items when surface forms refer to the same meaning.

Edges follow the KG schema defined in Section~\ref{sec:pkg-schema}. At the schema level, \texttt{(ProductType, has\_key, AttributeKey)} specifies which keys apply to a type, and \texttt{(AttributeKey, has\_value, Value)} assigns each value node to exactly one attribute key (the value typing constraint). At the instance level, \texttt{(Product, of\_type, ProductType)} links an item to its canonical product type, and \texttt{(Product, has\_attribute, Value)} encodes item-specific attribute
assertions.

The example can be read as follows. Items typed as \emph{Desktop Computer} connect to type-relevant keys such as \emph{GPU} and \emph{RAM Capacity}; those keys in turn type their value nodes (e.g., \emph{NVIDIA GeForce} for \emph{GPU}, \emph{8GB}/\emph{16GB} for \emph{RAM Capacity}). Item-level edges then link specific products to the values supported by their evidence (title/description/specifications/images). This design enables type checking (preventing values whose key is not defined for the product type) and canonicalization (merging synonymous surface variants into shared nodes), which is critical for continual KG growth under noisy marketplace data.

\section{WKE Weighting and Sensitivity}
\label{app:wke-weighting}

We use a 3:1:1 weighting in WKE to prioritize semantic validity. Acceptance is
human-labeled and directly measures whether induced types/keys are correct; in
contrast, Compression/Coverage (and P-Prec/P-Rec for keys) can be improved by
policies that change ontology size without improving correctness (e.g., aggressive
\texttt{MERGE} decisions). In a continually written KG, invalid schema nodes are
high-cost because they persist and can misguide retrieval and downstream extraction,
whereas lower compression or coverage is typically recoverable in later updates.

We performed an offline sensitivity check with alternative weights (2:1:1 and 4:1:1).
Using 2:1:1 occasionally elevated models/policies that achieved higher compression
primarily via over-merging, despite lower human acceptance. Using 4:1:1 produced
similar rankings to 3:1:1 but reduced the influence of efficiency/coverage. Overall,
3:1:1 best aligned the WKE ranking with our qualitative assessment and human
annotations by favoring models that are both accurate and reasonably efficient.

\section{AutoPKG Workflow Design}
\label{app:Multi-Agent Workflow}

\begin{figure*}[t]
\centering
\includegraphics[width=\textwidth]{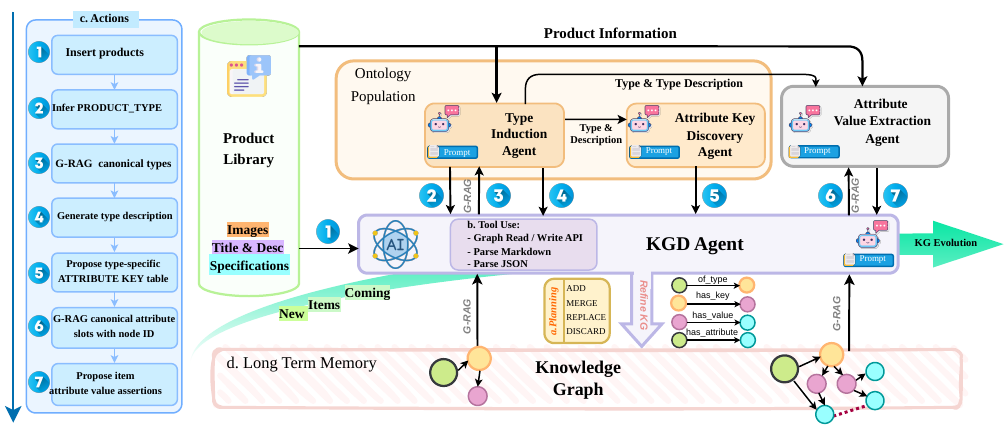}
\caption{Multi-Agent Framework.}
\label{fig:MA}
\end{figure*}

To support automatic and dynamic product-KG construction, AutoPKG adopts a modular, centralized multi-agent orchestration mechanism that delegates core subtasks—ontology population, attribute–value extraction, KG update, and optional refinement procedures—to specialized Agents, as illustrated in Figure~\ref{fig:MA} \citep{guo2024largelanguagemodelbased,lu2025karmaleveragingmultiagentllms, shi-etal-2025-answering, shi2026agentselectbenchmarknarrativequerytoagent}. 

Formally, agent communication follows a centralized protocol. Given a product input $x$,
each task-specific agent $i$ produces an intermediate output
\begin{equation}
a_i = \pi_i(h_i, x),
\end{equation}
where $\pi_i$ denotes the agent-specific policy implemented by an LLM, and $h_i$
represents the agent’s internal state. All agent outputs are transmitted to the
KGD agent, which aggregates them and determines the subsequent
KG operations.

Each agent operates as an autonomous functional module while coordinating through the KGD agent, which serves as the sole read–write interface to the KG. The Ontology Population agent block and the Attribute Value Extraction Agent performs as described in Sections~\ref{sec:ontology}, \ref{sec:pave}, \ref{sec:kgd-agent}, respectively.

The Ontology Population agent block performs a two-stage schema alignment process (product-type induction with canonical type-description generation, and key-attribute discovery) and hands the resulting schema to the Decision Agent for controlled integration. 
The Attribute Value Extraction Agent retrieves type-specific attribute tables via the Decision Agent and extracts attribute values under a RAG-style workflow, proposing updates to the KGD Agent when necessary.

The overall objective of the system is to incrementally construct a consistent and
saturated KG, which can be expressed as
\begin{equation}
\hat{G} = \arg\max_{G} R(G),
\end{equation}
where $R(G)$ measures KG completeness and structural consistency under the saturation
criterion.

\paragraph{Long-term memory and retrieval-coupled generation}
From a multi-agent perspective, we treat $\mathcal{G}$ as a long-term memory that stores canonicalized types/keys/values and their local structural neighborhoods. At each step, G-RAG retrieves context from $\mathcal{G}$ (e.g., nearest canonical nodes and type-conditioned key tables) to condition proposal generation and to constrain subsequent decisions. As a result, the pipeline is inherently \emph{non-linear}: each accepted update changes the retrieval landscape, which in turn reshapes future proposal distributions and progressively stabilizes the induced schema.

\paragraph{KGD: A planner with a Constrained Action Space}
While multiple agents contribute distinct competencies, AutoPKG centers on a \emph{Knowledge Graph Decision} agent (KGD) as the critical planner and \emph{sole write interface} to $\mathcal{G}$. Rather than directly committing open-ended model outputs, KGD resolves every upstream proposal through a constrained edit space---\texttt{ADD}, \texttt{MERGE}, \texttt{REPLACE}, or \texttt{DISCARD}---thereby projecting unconstrained generation into an explicit, auditable update language. This centralization makes continual canonicalization operational: KGD arbitrates deduplication, synonym consolidation, and label promotion using retrieved local neighborhoods as decision context, ensuring that graph growth remains controlled as coverage expands.

\paragraph{Tool-grounded execution and emergent bootstrapping.}
KGD is tool-grounded: it performs graph read and write via explicit APIs and parses structured LLM outputs (e.g., JSON/Markdown) into executable KG transactions (node/edge creation and insertion statements). The resulting retrieval--decision--write feedback loop enables \emph{self-bootstrapping} and emergent behavior: schema elements induced early become anchors for later extraction, repeated consolidation induces attractor-like canonical nodes, and the system continuously refines both ontology (types/keys) and facts (values) \textbf{without manual curation}. In aggregate, AutoPKG realizes a fully automated, continually updating PKG construction process whose reliability derives from the synergy between specialized proposal agents and KGD’s constrained, memory-conditioned planning.

\section{Human Annotation Protocol and Guidelines}
\label{app:human-annotation}

We use trained in-house annotators to evaluate (i) product type validity, (ii) attribute key validity, and
(iii) item--value edge correctness after KGD canonicalization. Annotation is performed as paid work by
professional annotators as part of their employment. The released dataset subset described in
Section~\ref{sec:dataset} has been approved for release by the company under internal governance.

\subsection{Annotation Materials and Setup}
Annotators are shown a single product listing consisting of a title, description/highlights, structured
specifications when available, and up to ten images. They are also shown the specific model output to be
judged. Annotators are instructed to base decisions only on the provided listing evidence and not to use
external resources.

\subsection{Labels}
For product type and attribute key evaluation, annotators assign Accept or Reject. For attribute value
evaluation, annotators assign Correct or Incorrect to each instantiated edge of the form
\texttt{(Product, has\_attribute, Value)} after KGD canonicalization.

\subsection{General Guidelines}
Annotators judge semantic correctness rather than surface form. Spelling variants, formatting differences,
and clear synonyms are acceptable (e.g., ``Gray'' vs.\ ``Grey''). When evidence is insufficient, annotators
should not guess and instead choose Reject or Incorrect. Outputs that are subjective or promotional in
nature (e.g., ``High Quality'', ``Best'', ``Premium'') are labeled Reject or Incorrect. When different fields
conflict (e.g., title vs.\ specifications), annotators label Incorrect unless one source clearly dominates
through repeated mentions, internal consistency, and/or strong visual support.

\subsection{Task 1: Product Type Validity}
A product type should describe what the item is, using a canonical category name that generalizes across
sellers. A prediction is labeled Accept if it is a valid, unambiguous product type supported by the listing,
and if it does not include brand names, model identifiers, marketing language, or attribute phrases. A
prediction is labeled Reject if it is incorrect for the listing, too vague to be actionable (e.g., ``Item'',
``Accessory''), primarily an attribute value (e.g., ``Red''), or an over-specific seller phrasing that does not
generalize.

\subsection{Task 2: Attribute Key Validity}
An attribute key should represent an inherent, type-relevant characteristic that a buyer can use to compare
products, rather than logistics or commercial metadata. A key is labeled Accept if it is applicable to the
product type, refers to a stable product characteristic (e.g., brand, material, capacity, compatibility), and
is not a redundant paraphrase of another key at the same meaning. A key is labeled Reject if it corresponds
to shipping, seller, pricing, warranty administration, SEO/marketing claims, or if it is too vague or
inapplicable to the type.

\subsection{Task 3: Attribute Value Edge Correctness}
Each predicted assertion is evaluated at the KG edge level after canonicalization. An edge is labeled
Correct only if the value is supported by evidence in the title, description, specifications, and/or images,
and if the value matches the intended attribute meaning under its key type after canonicalization. An edge
is labeled Incorrect if the value is not supported, is only weakly inferred, conflicts with stronger evidence,
or is attached to the wrong key type (for example, a battery form factor being recorded under a brand key).

\subsection{Sampling and Annotation Volume}
For product type validity, we sample 500 products and evaluate per-product type predictions from 8 models,
yielding 4{,}000 judgments. For attribute key validity, we sample a set of 500 product types, each model generates a
type-specific key table and we uniformly sample one key per product type per model, again yielding 4{,}000
judgments. For attribute value edges, annotators label instantiated edges
\texttt{(Product, has\_attribute, Value)} as Correct or Incorrect using the full listing evidence of the same set, containing 9{,}040 judgments.

\subsection{Recruitment and Payment}
Annotators are full-time professional annotators employed by the organization operating the marketplace. Annotation is conducted as part of paid work at rates aligned with local regulations and internal compensation standards.

\subsection{Consent and Data Use}
The annotated product listings are seller-provided catalog content. Annotators are instructed to use only the provided listing materials and not to access external resources. The study does not involve interaction with end users.



\section{LLM Judge Agreement for KGD Evaluation}
\label{app:judge-agreement}


\begin{table*}[t]
\centering
\small
\setlength{\tabcolsep}{6pt}
\begin{tabular}{lccccccc}
\toprule
& \textbf{GPT-5} & \textbf{Gemini-2.5-Pro} & \textbf{Qwen3-Max} & \textbf{Kimi-K2} & \textbf{DeepSeek-V3.1} & \textbf{Consensus} & \textbf{Human} \\
\midrule
\textbf{GPT-5}           & 1.000 & \textbf{0.812} & 0.687 & 0.695 & 0.573 & 0.798 & 0.799 \\
\textbf{Gemini-2.5-Pro}  & \textbf{0.812} & 1.000 & 0.710 & 0.730 & 0.580 & 0.827 & 0.794 \\
\textbf{Qwen3-Max}       & 0.687 & 0.710 & 1.000 & \textbf{0.812} & \textbf{0.730} & 0.861 & 0.764 \\
\textbf{Kimi-K2}         & 0.695 & 0.730 & \textbf{0.812} & 1.000 & 0.696 & \textbf{0.876} & 0.750 \\
\textbf{DeepSeek-V3.1}   & 0.573 & 0.580 & \textbf{0.730} & 0.696 & 1.000 & 0.721 & 0.635 \\
\textbf{Consensus}       & 0.798 & 0.827 & 0.861 & \textbf{0.876} & 0.721 & 1.000 & \textbf{0.837} \\
\textbf{Human}           & 0.799 & 0.794 & 0.764 & 0.750 & 0.635 & \textbf{0.837} & 1.000 \\
\bottomrule
\end{tabular}
\caption{Pairwise Cohen’s $\kappa$ agreement among the LLM judges and human annotators used in Section~\ref{sec:kgd-eval}.}
\label{tab:llm-judge-kappa}
\end{table*}



In Section~\ref{sec:kgd-eval}, we evaluate candidate KGD backbones against a reference label derived from majority vote among five frontier LLM judges. To assess the reliability of this reference—and to justify its use as a scalable evaluator for KGD edit actions—we analyze inter-judge agreement, including alignment with human annotations. Specifically, we compute pairwise Cohen’s $\kappa$ for all judge pairs, including human annotators on sample of 2,000 KGD decision instances (10,000 total judgments from 5 AI judges), on the same set of KGD decision instances used to form the 3/5 consensus labels. Cohen’s $\kappa$ quantifies agreement beyond chance for categorical decisions, with higher values indicating greater consistency. As shown in Table~\ref{tab:llm-judge-kappa}, the LLM judges exhibit substantial mutual agreement, and—critically—their consensus aligns closely with human judgments (e.g., $\kappa = 0.837$ between the Consensus panel and Human). This strong alignment supports the use of majority-vote LLM consensus as a reliable and scalable surrogate for human evaluation in KGD edit assessment.

\section{Open-Source Dataset Comparison Experimental Setup}
\label{app:Open-Dataset-Experimental}

\begin{table*}[t]
\centering
\footnotesize
\setlength{\tabcolsep}{3pt}
\renewcommand{\arraystretch}{1.15}

\begin{tabularx}{\textwidth}{@{}
p{1.6cm}  
p{2.6cm}  
c         
c         
c         
c         
p{1.2cm}  
c         
>{\raggedright\arraybackslash}X 
@{}}
\toprule
\textbf{Dataset} &
\makecell{\textbf{Release}\\\textbf{Year}} &
\makecell{\textbf{\#}\\\textbf{Products}} &
\makecell{\textbf{\#}\\\textbf{Categories}} &
\makecell{\textbf{Avg.\ \#Keys}\\\textbf{per Product}} &
\makecell{\textbf{Avg.\ \#Values}\\\textbf{per Key}} &
\makecell{\textbf{Lan-}\\\textbf{guage}} &
\makecell{\textbf{Multi-}\\\textbf{modal}} &
\makecell[l]{\textbf{Applicable Fields}\\\textbf{\& Introduction}} \\
\midrule
ImplicitAVE & ACL 2024 Findings & 70{,}214  & 167 & 1.000 & 1.000 & English & \cmark &
Sampled from MAVE. \\
MAVE & WSDM 2022 & 275{,}049 & 950 & 1.629 & 3.395 & English & \xmark &
Amazon. \\
AE--650k & ACL 2019 & 39{,}505 & 558 & 2.250 & 1.000 & English & \xmark &
AliExpress. \\
\bottomrule
\end{tabularx}

\caption{Open-source dataset statistics.}
\label{tab:dataset_comparison}
\end{table*}

We adopt widely used public AVE task datasets as benchmarks to ensure transparent, reproducible comparison and to validate generalization beyond any single marketplace snapshot. Concretely, we construct evaluation splits by sampling high-quality product–attribute instances from three influential datasets that collectively cover both multimodal and text-only listing regimes—ImplicitAVE (text+images), MAVE (text), and AE-650K (text)—thereby spanning diverse categories, attribute taxonomies, and annotation styles. These benchmarks have broad community uptake and are among the most frequently cited product-attribute resources in recent years, which makes them a strong reference point for positioning performance. The resulting evaluation set sizes (70,214 / 275,049 / 39,505 instances, respectively) provide substantial coverage while keeping the protocol consistent across settings. Table~\ref{tab:dataset_comparison} lists the details of these datasets.

\subsection{Fine-grained KG Node Quality Comparison Experiment}
\label{app:AP-KG-Node-Quality-Comparison}

\begin{table*}[t]
\centering
\small
\setlength{\tabcolsep}{5pt}
\begin{tabular}{@{}l p{3.9cm} ccc ccc@{}}
\toprule
\multirow{2}{*}{Datasets} & \multirow{2}{*}{KG Node Quality Comparison} &
\multicolumn{3}{c}{Attribute Key Node} & \multicolumn{3}{c}{Attribute Value Node} \\
\cmidrule(lr){3-5}\cmidrule(lr){6-8}
& & P & R & F1 & P & R & F1 \\
\midrule

\multirow{3}{*}{ImplicitAVE}
& $\text{KG}_1$\textsubscript{\footnotesize\url{raw}}
& \textbf{0.8264} & 0.1028 & 0.1789
& 0.6945 & 0.0828 & 0.1449 \\
& $\text{KG}_2$\textsubscript{\footnotesize\url{google/gemma-3-27b}}
& 0.7289 & 0.6265 & 0.6641
& \textbf{0.9031} & 0.7854 & \textbf{0.8285} \\
& $\text{KG}_3$\textsubscript{\footnotesize\url{Qwen/Qwen3-VL-32B}}
& 0.6858 & \textbf{0.7132} & \textbf{0.6909}
& 0.7607 & \textbf{0.8014} & 0.7707 \\
\midrule

\multirow{3}{*}{Mave}
& $\text{KG}_1$\textsubscript{\footnotesize\url{raw}}
& 0.6339 & 0.1677 & 0.2458
& \textbf{0.7669} & 0.178 & 0.2736 \\
& $\text{KG}_2$\textsubscript{\footnotesize\url{google/gemma-3-27b}}
& \textbf{0.6908} & \textbf{0.4107} & \textbf{0.4909}
& 0.453 & 0.2472 & 0.3018 \\
& $\text{KG}_3$\textsubscript{\footnotesize\url{Qwen/Qwen3-VL-32B}}
& 0.1665 & 0.1569 & 0.1476
& 0.3653 & \textbf{0.3817} & \textbf{0.3471} \\
\midrule

\multirow{3}{*}{AE--650k}
& $\text{KG}_1$\textsubscript{\footnotesize\url{raw}}
& \textbf{0.8392} & 0.2997 & \textbf{0.4127}
& \textbf{0.8002} & 0.2423 & \textbf{0.3447} \\
& $\text{KG}_2$\textsubscript{\footnotesize\url{google/gemma-3-27b}}
& 0.4057 & 0.128 & 0.1793
& 0.6359 & 0.2399 & 0.325 \\
& $\text{KG}_3$\textsubscript{\footnotesize\url{Qwen/Qwen3-VL-32B}}
& 0.5138 & \textbf{0.3231} & 0.3633
& 0.4452 & \textbf{0.2611} & 0.2987 \\
\bottomrule
\end{tabular}
\caption{Fine-grained KG edge-level quality evaluation results on three datasets.}
\label{tab:kg_finegrained_node_quality}
\end{table*}
\paragraph{Rationale for Edge-level Precision and Recall}

We evaluate product–attribute KGs using edge-level precision and recall at both the attribute key and attribute value granularity. This choice reflects the core objectives of dynamic product KG construction and aligns with standard evaluation practices in knowledge extraction and KG population.

At the attribute key level, edge-level precision measures the correctness of induced attribute schemas, indicating whether predicted attribute nodes correspond to semantically valid and product-relevant attributes. Recall measures schema coverage, capturing whether salient attributes present in a reference KG are successfully recovered. Together, they quantify the quality of attribute induction and canonicalization.

At the attribute value level, edge-level metrics assess the correctness and completeness of atomic key–value facts, abstracting away from graph topology and focusing on factual content. High precision reflects accurate value extraction, while high recall indicates effective population of attribute values.

Crucially, edge-level performance also reflects graph construction quality beyond extraction accuracy. In AutoPKG, all updates are mediated by the centralized Knowledge Graph Decision (KGD) agent. High precision implies effective filtering of noisy proposals, while high recall indicates that consolidation and alignment decisions do not suppress valid knowledge. Thus, edge-level metrics jointly capture extraction quality and the effectiveness of knowledge fusion and canonicalization.

\paragraph{Data Subsets and Input Modalities}
To balance computational cost and dataset diversity, we evaluate on stratified subsets from three public benchmarks:

ImplicitAVE (multimodal): 10,000 products

MAVE (text-only): 55,000 products

AE-650k (text-only): 39,505 products

These subsets preserve original category distributions while remaining tractable for large-scale LLM-based KG construction. For text-only datasets, we use the concatenation of product titles and detailed descriptions as model input, reflecting realistic e-commerce settings where images may be unavailable.

\paragraph{Knowledge Graph Construction}

\begin{table*}[t]
\centering
\small
\setlength{\tabcolsep}{5pt}
\renewcommand{\arraystretch}{1.15}
\label{tab:kg-stats}
\resizebox{\textwidth}{!}{%
\begin{tabular}{llcccccccc}
\toprule
\multirow{2}{*}{Dataset} &
\multirow{2}{*}{kg version} &
\multicolumn{4}{c}{Node Size} &
\multicolumn{4}{c}{Edge Size} \\
\cmidrule(lr){3-6}\cmidrule(lr){7-10}
& & Product & Type & Key & Value & Has Type & Has Key & Has Value & Has Attribute \\
\midrule

\multirow{3}{*}{ImplicitAVE}
  & KG\_1\textsubscript{\footnotesize\url{raw}}    & 10000 & --  & 23   & 152  & --    & 10000 & 152  & -- \\
  & KG\_2\textsubscript{\footnotesize\url{gemma}}  & 10000 & 779 & 1178 & 2604 & 10000 & 9561  & 2612 & 51049 \\
  & KG\_3\textsubscript{\footnotesize\url{Qwen}}   & 10000 & 779 & 1178 & 3692 & 10000 & 9561  & 3703 & 106680 \\
\midrule

\multirow{3}{*}{MAVE}
  & KG\_1\textsubscript{\footnotesize\url{raw}}    & 55010 & --   & 491  & 16379 & --    & 55010 & 16379 & -- \\
  & KG\_2\textsubscript{\footnotesize\url{gemma}}  & 55010 & 1462 & 2780 & 15181 & 55010 & 19241 & 15183 & 175656 \\
  & KG\_3\textsubscript{\footnotesize\url{Qwen}}    & 55010 & 1462 & 2780 & 18498 & 55010 & 19241 & 18498 & 330545 \\
\midrule

\multirow{3}{*}{AE--650k}
  & KG\_1\textsubscript{\footnotesize\url{raw}}    & 39505 & --  & 2584 & 12497    & --    & 39505 & 12497 & -- \\
  & KG\_2\textsubscript{\footnotesize\url{gemma}}  & 39505 & 558 & 1344 & 2652 & 39505 & 7013  & 2652  & 83076 \\
  & KG\_3\textsubscript{\footnotesize\url{Qwen}}    & 39505 & 558 & 1344 & 3530 & 39505 & 7013  & 3530  & 177849 \\
\bottomrule
\end{tabular}%
}

\caption{Structural statistics of KGs constructed under different variants. 
For KG$_{\text{raw}}$, type nodes and the corresponding \texttt{has\_attribute} and \texttt{of\_type} relations are intentionally omitted, as this design represents a deliberate trade-off between storage footprint and computational efficiency.}

\label{tab:Statistics-of-node/edge}
\end{table*}
We construct three KGs from the same product subsets:

KG\textsubscript{raw}.
A baseline KG built directly from dataset triplets. The graph is implemented in Neo4j with global node deduplication. Two relation types are used: \textsc{Has\_Key} (product–attribute) and \textsc{Has\_Value} (attribute–value).

KG\textsubscript{Gemma} / KG\textsubscript{Qwen-VL}.
Two predicted KGs constructed using the full AutoPKG framework, starting from an empty graph. All schema induction, extraction, and updates are proposed by agents and executed exclusively through the KGD agent. The two variants differ in the Multimodal Attribute Value Extraction model: isolating model effects from graph-level consolidation.

Structural statistics of KGs constructed under different variants are reported in Table~\ref{tab:Statistics-of-node/edge}.

\begin{figure*}[t]
\centering
\begin{casepromptbox}{KG ground-truth construction prompt}
\small
Based on Product Description, extract relevant attribute name and its corresponding attribute value of this product, and Refer to the Hypotheses list when providing the answer. Be very accurate!

A. Task Description:

You are given \textbf{2} information sources. 1: Original Product Description; 2. A \textbf{Hypotheses list of attributes and specifications under review}, it is provided for you to review and judge after the Step1, at the Step2.

\textbf{Step1}: You should first independently determine the attributes of this product in an e-commerce context based on the product Description, providing this information to both buyers and sellers.
Then, you should fill in the values for these attributes based on the product information. 

- Do not rely on *\textbf{Hypotheses list of attributes and specifications under review*} at this step, as they may be inaccurate and incomplete.

- Analyze only based on the Description in Step1.

\textbf{Step2}: Based on the \textbf{Hypotheses list of attributes and specifications under review} and the names/values you determained on the Step1:

- If the existing attribute names and values are correct → Adopt them directly. Using the same words for correct name/values in Hypotheses list!

- If they are inaccurate → Correct them.

- If they are incomplete → Add new ones.

\textbf{Step3}: Output Only in Markdown table format like:

For a prodcut Pen:

| Attribute Name         | Value                    |

| ---------------------- | ------------------------ |

| Writing Mechanism      | Retractable              |

| Ink Color              | Black                    |

| etc.                   | etc.                     |

B. Criterias:

The \textbf{attributes names} must meet these criterias:
- Focused on \textbf{inherent product characteristics} (not logistics, packaging, or SEO)
- Based on \textbf{industry standards or common e-commerce practices}
- Includes \textbf{visual and functional attributes} that influence purchasing decisions

The corresponding \textbf{attributes values} meet these criterias:

- For attributes that have multiple values, return a list of values.

C. Provided information:

1. Real Product Image and the Product Description you need to use:

Product Description: 
[Title] Tommy Hilfiger Big Boys' Short Sleeve 85  Shirt
Product image 1 (if have): 
\begin{center}
\includegraphics[width=0.15\textwidth]{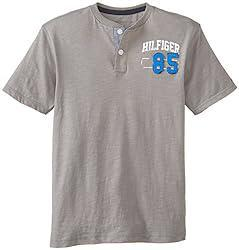}
\end{center}

2. \textbf{Hypotheses list of attributes and specifications under review}, it is provided for your reference:

Attribute name 1:

Neckline

Attribute value 1:

Henley

Rest of attributes:

- Brand (71009): Tommy Hilfiger

- Care Instructions (71057): Machine Washable

- Closure Type (71018): Button

- Color (71006): Gray

- Drape (71176): Soft

- Dress Length (71446): Tunic

- Elasticity (71071): Slight Stretch

- Fit Type (71095): Regular Fit

- Material (71010): Cotton

- Size (71007): 85

- Sleeve Style (71049): Short Sleeve
(truncated for brevity)

\hrulefill

\vspace{0.5em}
\textbf{Expected Output:}

\{
color:[
0:"grey"
],
brand:"Tommy Hilfiger",
material:"cotton",
closure type:"button",
pattern:"solid",
sleeve style:"short sleeve",
neckline:"Henley",
fit type:"fitted",
lining material:"unlined"
\}
\end{casepromptbox}
\caption{Prompt for KG ground-truth construction. Optional visual inputs are used only for multimodal tasks. Both raw and AutoPKG-generated KGs are normalized into a unified node--edge schema, after which LLM outputs are parsed into triples and integrated into the KG by the KGD.}

\label{fig:prompt-gt-example}
\end{figure*}

\paragraph{Reference KG Construction}
As no dataset provides a fully canonicalized product KG, we construct a reference graph (KG*) using a G-RAG-based annotation pipeline with the Qwen3-VL-235B LLM annotator. The prompting enforces canonical attribute naming, value normalization, and duplicate suppression, yielding a globally consistent \textbf{reference KG} as our silver level ground truth. Example prompts and outputs are shown in Figure~\ref{fig:prompt-gt-example}

\paragraph{Calculation Method}
For each product $i$, let $K_i$ and $\hat{K}_i$ denote the sets of distinct attribute keys in $\mathrm{KG}^{\star}$ and a predicted KG, respectively, and let $V_i$ and $\hat{V}_i$ denote the sets of distinct key--value facts $(k,v)$.

\noindent\textbf{Key-level scoring}
We compute per-product precision and recall as
\[
P_K(i)=\frac{\lvert \hat{K}_i \cap K_i \rvert}{\lvert \hat{K}_i \rvert},
\qquad
R_K(i)=\frac{\lvert \hat{K}_i \cap K_i \rvert}{\lvert K_i \rvert}.
\]

\noindent\textbf{Value-level scoring}
Value true positives require key agreement and a substring match:
\[
(k,v)\in \mathrm{TP}_V(i)
\iff
\exists (k,v^{\star})\in V_i \ \text{s.t.}\ v^{\star} \supseteq v,
\]
yielding
\[
P_V(i)=\frac{\lvert \mathrm{TP}_V(i)\rvert}{\lvert \hat{V}_i \rvert},
\qquad
R_V(i)=\frac{\lvert \mathrm{TP}_V(i)\rvert}{\lvert V_i \rvert}.
\]

\paragraph{Results Explained}
We report macro-averaged precision, recall, and F1 over the intersection of predicted and reference product IDs, and report the results in Table~\ref{tab:kg_finegrained_node_quality}.

Across three datasets (ImplicitAVE, MAVE, and AE-650K), our LLM-constructed PKG (\(\mathrm{KG}_2\)) improves exact-match macro-\(F_1\) over the strongest baseline \(\mathrm{KG}_1\) (raw triplet KG directly converted from the original product -> attribute name and attribute -> attribute value data.), with \texttt{google/gemma-3-27b-it} delivering the largest average gains: \(+0.166\) on \textsc{AttributeKey} nodes and \(+0.231\) on \textsc{AttributeValue} nodes across ImplicitAVE/MAVE/AE--650k (absolute macro-\(F_1\), averaged across datasets), up to \(+0.684\) F1 gain on multimodal ImplicitAVE. The improvements are most pronounced on the multimodal ImplicitAVE setting, where \(\mathrm{KG}_2\) substantially increases both key and value \(F_1\) relative to \(\mathrm{KG}_1\), indicating markedly better node coverage under exact matching. On MAVE, \(\mathrm{KG}_2\) increases key-level \(F_1\) and yields modest but consistent gains on value nodes, suggesting that graph-guided consolidation reduces missing and redundant schema elements. On AE-650k, key-node \(F_1\) can drop for some extractors, while value-node \(F_1\) remains close to the baseline, reflecting a harder consolidation regime with stricter canonicalization requirements. Overall, the gains align with the G-RAG-style decision mechanism in KGD: retrieving the local KG neighborhood during edit validation provides schema context that suppresses spurious node additions and improves recall through consistency-aware merging.

\subsection{Downstream PAVE Task Assessment}
\label{app:AP-PAVE-Task-Assessment}
G-RAG has been demonstrated to be an effective paradigm for enhancing the generative capabilities of large language models (LLMs). 
Within the AutoPKG framework, the methodological design presented in Section~\ref{sec:pave}, together with the experimental results reported in Section~\ref{sec:results-value}, empirically demonstrates the performance gains of G-RAG in attribute value extraction tasks.

Following the same implementation protocol described in Figure~\ref{fig:AutoPKG} and Section~\ref{sec:pave}, we extend the framework to realize G-RAG in three external datasets as illustrated in Appendix~\ref{app:Open-Dataset-Experimental}, ensuring a fair and controlled comparison across retrieval paradigms. 
Let \(TP_i=\{(k,v)\in \hat{Y}_i:\exists (k,v^\star)\in Y_i \text{ s.t. } v^\star \supseteq v\}\), requiring key agreement and substring value matching. Per-product precision and recall are
\[
P(i)=\frac{|TP_i|}{|\hat{Y}_i|},\qquad R(i)=\frac{|TP_i|}{|Y_i|}.
\]
Below, we illustrate how retrieved subgraph content applied to the PAVE generative task, showcasing how graph-structured knowledge can be effectively integrated into the generation process.

\paragraph{G-RAG for Attribute Keys}
The product type and the candidate attributes, together with their descriptions, examples, and synonyms, are retrieved as a text-based subgraph from the existing KG and provided as reference context and candidate guidance for attribute-value generation for a specific product. This reference content is incorporated into the prompt for the generative LLM in the same manner as in Figure~\ref{fig:prompt-attr-key}

\paragraph{G-RAG for Attribute Values}
Type, Attribute Candidates with Description, Examples and Synonyms are retrieved as a text-based sub-graph from the existing KG and provided as a reference and candidates  for potential attribute value generation, for a specific product. The reference content if put into the prompt for generative LLM in the same way as Figure~\ref{fig:prompt-attr-value-example}

\paragraph{Results}
We report macro-averaged \(P,R,F_1\) over product IDs in Table~\ref{tab:ave-main}. All runs are inference-only; the system input is \((x_i,\mathcal{G})\) and the output is a deduplicated set of \((pid,k,v)\) facts.





\section{Human Validation of LLM-based Annotations for Sections~\ref{sec:public-eval-kg} and~\ref{sec:public-eval-pave}}
\label{app:judge-agreement-open}

\begin{table*}[t]
\centering
\small
\begin{tabular}{lcccccccc}
\toprule
& \multicolumn{4}{c}{\textbf{Section~\ref{sec:public-eval-kg}: Edge-Level Assessment}} 
& \multicolumn{4}{c}{\textbf{Section~\ref{sec:public-eval-pave}: PKG-Augmented PAVE}} \\
\cmidrule(lr){2-5} \cmidrule(lr){6-9}
\textbf{Dataset} 
& \multicolumn{2}{c}{KG Attribute Key Nodes} 
& \multicolumn{2}{c}{KG Attribute Value Nodes}
& \multicolumn{2}{c}{Attribute Key} 
& \multicolumn{2}{c}{Attribute Value} \\
\cmidrule(lr){2-3} \cmidrule(lr){4-5}
\cmidrule(lr){6-7} \cmidrule(lr){8-9}
& $\kappa$ & Agree (\%) & $\kappa$ & Agree (\%) 
& $\kappa$ & Agree (\%) & $\kappa$ & Agree (\%) \\
\midrule
ImplicitAVE 
& 0.8251 & 86.41 & 0.8015 & 84.32 
& 0.8572 & 85.90 & 0.7871 & 80.18 \\
MAVE 
& 0.8408 & 89.73 & 0.8392 & 84.69 
& 0.8429 & 87.12 & 0.8128 & 84.08 \\
AE650k 
& 0.8583 & 87.58 & 0.8135 & 87.52 
& 0.8515 & 86.66 & 0.8014 & 80.62 \\
\bottomrule
\end{tabular}

\caption{Agreement between LLM-based annotations and human labels on public datasets for Section~\ref{sec:public-eval-kg} (edge-level assessment) and Section~\ref{sec:public-eval-pave} (PKG-augmented PAVE). We report Cohen’s Kappa ($\kappa$) and exact agreement rate.}
\label{tab:llm-human-agreement}

\end{table*}

To assess the reliability of the LLM-generated annotations used for constructing silver reference KGs on public datasets, we conduct a human validation study on a randomly sampled subset of 1,000 instances drawn from ImplicitAVE, MAVE, and AE-650k. Human annotators independently annotate these instances following the same protocols as those used in the Section~\ref{sec:public-eval-kg} edge-level assessment and the Section~\ref{sec:public-eval-pave} PKG-augmented PAVE task.

We measure the agreement between LLM-based annotations (produced by Qwen3-VL-235B-A22B-Instruct) and human labels using Cohen’s Kappa and exact agreement rate. The results indicate strong and consistent alignment between LLM and human annotations across both tasks:

\begin{itemize}
    \item \textbf{KG Edge-Level Assessment (Section~\ref{sec:public-eval-kg}):} 
    
    Cohen’s Kappa = 0.8297, agreement rate = 86.71\%;
    \item \textbf{PKG-Augmented PAVE (Section~\ref{sec:public-eval-pave}):} 
    
    Cohen’s Kappa = 0.8255, agreement rate = 84.09\%.
\end{itemize}

These results indicate substantial agreement and support the reliability of the LLM-based annotations as a scalable proxy for human judgment in both edge-level KG evaluation and PKG-augmented reasoning tasks, enabling substantial reductions in annotation effort (approximately 2,560 annotator-hours).

\section{Data Release, Privacy, and Content Safety}
\label{app:data-release}

We plan to release, upon acceptance, a vetted subset of our product-listing sample together with a snapshot of the constructed product--attribute knowledge graph. The release will include (i) product listing text (title/description/specifications), (ii) associated product image URLs, and (iii) the PKG snapshot.

The data is derived from seller-provided catalog content and is not intended to identify individual people. It does not include end-user data (e.g., user profiles, messages, or behavioral logs). Nevertheless, to reduce privacy and safety risks prior to release, we will apply content-safety screening to exclude listings flagged for potentially unsafe content (e.g., adult or violent content). We will also remove or redact obvious personally identifying information (PII) patterns in text when present (e.g., phone numbers, email addresses, and messaging handles). Images will be screened using automated detectors, followed by manual spot checks on random samples to validate screening quality.

\section{Hardware and Inference Details}
\label{app:hardware}

\paragraph{Hardware.}
All experiments were run as containerized batch-inference jobs on Alibaba Cloud with NVIDIA H20 GPUs (141~GB GPU memory), using a CUDA 12.8 / PyTorch 2.8.0 runtime image (Python 3.10.13, GCC 13).

\paragraph{Serving stack.}
We use vLLM for inference when supported by the model and deployment environment; otherwise we fall back to standard HuggingFace inference with equivalent decoding settings. We enable prefix caching when available to reduce repeated prompt overhead in batched evaluation.

\paragraph{Inference details.}
All LLM/MLLM backbones are evaluated in a zero-shot setting using fixed prompt templates (Appendix~\ref{app:prompts}) and \emph{non-thinking} inference (i.e., without explicit test-time reasoning modes).
Unless otherwise noted, decoding hyperparameters are held constant across models: temperature $=0.7$, top-$p=0.8$, top-$k=20$, and max\_new\_tokens $=6400$.
For attribute value extraction, we include up to 10 images per product (\texttt{limit\_mm\_per\_prompt}$=10$).
We set GPU memory utilization to 0.9 for batch serving (\texttt{gpu\_memory\_utilization}$=0.9$).
For neighborhood context construction in Section \ref{sec:kgd-agent}, we set the retrieval hyperparameter $k = 10$.

\section{Configuration Suggestions}
\label{app:configuration_s}

The \textit{Minimal} config leverages smaller backbones to achieve the lowest cost but suffers in multimodal extraction fidelity (Edge $F_1$: 0.482). The \textit{Balanced} config upgrades only the value extractor; this yields a substantial quality leap (+3.4\% in Avg Quality) by maximizing Edge $F_1$ to 0.531---matching the Full config's extraction performance---for a negligible cost increase (+3.7\%). The \textit{Full} config further employs maximum-capacity models for KGD and Key Discovery, improving average quality by an additional +1.1\% (total +4.6\% over Minimal) at a moderate premium (+16.5\%). These results suggest that for most production scenarios, the \textit{Balanced} tier offers the optimal return on investment, securing high-fidelity attribute extraction without the overhead of larger schema induction models.

\section{Prompts Used in AutoPKG}
\label{app:prompts}

\subsection{KGD Prompt}
\label{app:prompt-kgd}

KGD is the sole write interface to the PKG. For each proposed node (or edge), KGD is provided with (i) a set of top-$k$ retrieved \emph{relevant KG nodes} of the same node type (e.g., \textsc{ProductType}, \textsc{AttributeKey}, \textsc{Value}), and (ii) a \emph{candidate} node generated by upstream agents (e.g., product type suggestion, schema induction, or value extraction). KGD then selects one of four actions---\texttt{ADD}, \texttt{MERGE}, \texttt{REPLACE}, or \texttt{DISCARD}---to ensure global consistency and prevent redundancy. In particular, \texttt{MERGE} collapses surface-form variants into an existing canonical node, \texttt{REPLACE} updates the canonical label to a more standard naming, and \texttt{DISCARD} filters vague or weakly grounded proposals. To enable lightweight post-processing, the output is constrained to a single-line decision (action only, or action plus matched \texttt{node\_id}).

Figure~\ref{fig:prompt-kgd-example-basic} shows an example prompt instance for product type canonicalization, where the candidate \emph{Wall Anchor} is compared against retrieved neighbors such as \emph{Wall Anchors} and \emph{Concrete Anchor} to decide whether to add a new type or merge with an existing one.
Figure~\ref{fig:prompt-kgd-example-strict}, ~\ref{fig:prompt-kgd-example-no-discard} shows further versions.

\subsection{Product Type Induction}
\label{app:prompt-type}

Figure~\ref{fig:prompt-product-type} shows a lightweight prompt to normalize each raw product listing into a \emph{canonical product type} string. The prompt instructs the model to remove non-type information (e.g., brand names, model numbers, attribute phrases, and marketing language) and to abstract the input to the most general category that remains unambiguous. This design encourages consistent, reusable type nodes (improving mergeability across sellers and countries) while avoiding over-generalization that would harm downstream attribute schema induction. If the input is not a valid product or lacks sufficient evidence, the model is required to abstain by returning \texttt{None}. The resulting type proposal is then passed to KGD (Section~\ref{sec:kgd-agent}) for graph insertion or canonicalization via \texttt{ADD}/\texttt{MERGE}/\texttt{REPLACE}/\texttt{DISCARD}.

\subsection{Attribute Key Discovery}
\label{app:prompt-attr-key}

Figure~\ref{fig:prompt-attr-key} shows the prompt used to generate a type-specific
attribute-key table (standard keys, definitions, and representative value examples)
conditioned on the inferred product type and its short type description. The model
must output \emph{only} a Markdown table, with attributes sorted by buyer importance,
starting with \texttt{Brand}. Values in the \texttt{Examples} column are constrained
to proper title case and 5--10 representative examples when applicable.

\subsection{Attribute Value Extraction}
\label{app:prompt-attr-value}

\begin{figure*}[t]
\centering
\begin{casepromptbox}{Attribute value extraction prompt example (multimodal). }
\small
\textbf{Instruction:}
Given the table of attributes as a reference, extract relevant attributes from the provided input text and image. Return a JSON object containing only the attribute ids and their corresponding attribute values.

\vspace{-1em}
\begin{itemize} \setlength\itemsep{-0.7em}
\item Extract JSON of attribute id and its corresponding attribute value.
\item For attributes that have multiple values, return a list of values.
\item If an attribute is not mentioned or cannot be determined with confidence, return null for that attribute.
\item You are not restricted to the specific values in the example table — if another value makes sense based on the input, feel free to use it.
\item Cross-validate conflicting or ambiguous information between the text and image (e.g., if the text says “100\% cotton” but the image shows a shiny fabric, consider whether it might be polyester or a blend).
\item Resolve ambiguities or duplicate values by selecting the most appropriate one based on context and consistency (e.g., choosing between "L" and "Large" based on other entries or general conventions).
\item Ensure the JSON output follows the structure and naming conventions shown in the example attribute table.
\item Provide only the JSON, no extra text or explanation.
\end{itemize}
\vspace{-1em}

\textbf{Simple Example:}
\vspace{-1em}
\begin{itemize} \setlength\itemsep{-0.7em}
\item Input text: Brand: Philips. The machine body is made of ABS plastic and is available in White, Black, and Silver. It has an IP44 rating for basic dust and splash protection.
\item Output JSON: \{"123": "Philips", "124": "IP44", "125": "ABS Plastic", "126": ["White", "Black", "Silver"]\}
\end{itemize}
\vspace{-1em}

\textbf{Product Type:} Battery Holder

\vspace{0.25em}
\textbf{Product Type Description:}
A product type designed to securely hold and connect batteries within electronic devices, providing electrical contact and physical support for standard battery sizes such as AA, AAA, or coin cells. Commonly used in consumer electronics, remote controls, and portable devices.

\vspace{0.5em}
| Attribute ID | Attribute Name | Description | Examples |
\\
|--------------|----------------|-------------|----------|
\\
| 18392 | Brand | The Manufacturer Or Company That Produces The Battery Holder | Duracell, Energizer, Panasonic, Anker, Varta |
\\
| 18597 | Battery Type | The Type Of Battery The Holder Is Designed To Accommodate | AA, AAA, C, D, 9V |
\\
| 19903 | Number of Batteries Supported | The Total Number Of Batteries The Holder Can Support | 1, 2, 3, 4, 6 |
\\
(truncated for brevity)

\vspace{0.5em}
\textbf{Title:} DIY 6-Pin Battery Holder for 3x 18650 Lithium-Ion Batteries

\vspace{0.5em}
\textbf{Highlight:} Compact and lightweight design for easy installation and portability.  - Dimensions: 78 x 60 x 21 mm / 3.1\" x 2.4\" x 0.83\" (L*W*H)  - Weight: 24g  - Material: Hard plastic and metal  - Color: Black  Package Include:  1 x 6-Pin Battery Holder for 18650 Batteries

\vspace{0.5em}
\textbf{Description:} Features: 1. This DIY 6-pin battery holder is designed for 3x 18650 lithium-ion batteries (batteries not included). 2. Made from durable hard plastic and metal, ensuring long-lasting use. 3. Compact and lightweight design for easy installation and portability. Specification: - Dimensions: 78 x 60 x 21 mm / 3.1" x 2.4" x 0.83" (L*W*H) - Weight: 24g - Material: Hard plastic and metal - Color: Black Package Include: 1 x 6-Pin Battery Holder for 18650 Batteries

\vspace{0.5em}
\textbf{Specifications:} \{"batteries\_type": ["D"], "warranty\_type": ["No Warranty"], "Plug\_Type": ["Universal"]\}

\begin{center}
\includegraphics[width=0.2\textwidth]{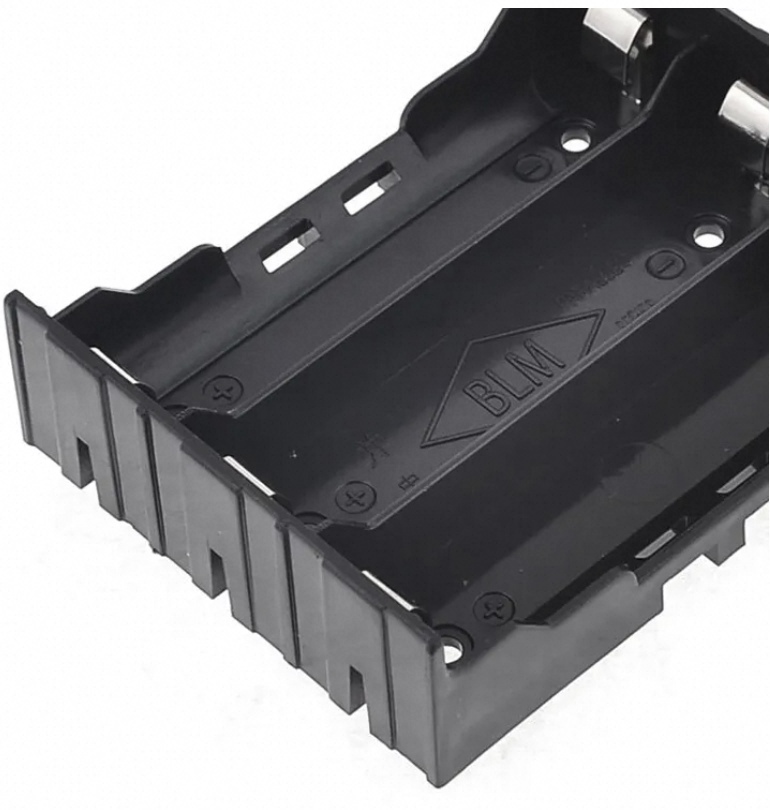}
\end{center}

\hrulefill

\vspace{0.5em}
\textbf{Expected Output:}

\{
"18392": "BLM",
"18597": "18650",
"19903": 3,
"27091": "Series",
"18809": "Nickel-Plated Brass",
"18407": "Snap-In",
"20066": "Positive Front",
"18427": "-20°C to 70°C",
"18724": null,
"19542": "ABS Plastic"
\}

\end{casepromptbox}
\caption{Attribute value extraction prompt example (multimodal). Given a product type and its attribute table (with IDs), the model extracts values from the listing text and image and returns a JSON object keyed by attribute ID, using lists for multi-valued fields and \texttt{null} when evidence is insufficient.}
\label{fig:prompt-attr-value-example}
\end{figure*}

Figure~\ref{fig:prompt-attr-value-example} presents the multimodal prompt used for attribute value extraction once a product type has been determined and its attribute schema (IDs, names, descriptions, and example values) is available. The model is instructed to read the product title, highlights, description, structured specifications, and the associated image, and to populate a JSON dictionary keyed \emph{only} by attribute IDs. To support downstream deterministic ingestion, the output format is tightly constrained: attributes with multiple mentions are returned as lists, and any attribute that is not explicitly stated or cannot be inferred with high confidence is set to \texttt{null}.

\begin{figure*}[t]
\centering
\begin{casepromptbox}{KGD prompt example (Basic Version)}
\small
\textbf{Instruction:}
You are a product knowledge graph decision agent.

\vspace{0.5em}
\textbf{Relevant nodes from KG:}

[\{'node\_id': 4587, 'node\_name': 'Wall Anchors', 'description': 'A category of fastening devices designed to securely attach objects to hollow or brittle walls ...'\}, 
\{'node\_id': 3762, 'node\_name': 'Tie-Down Anchor', 'description': 'A specialized hardware component ...', 'synonyms': ['Tie-down loops']\}, 
... (truncated) ...]

\vspace{0.5em}
\textbf{Node Type:} Product Type

\vspace{0.5em}
\textbf{Candidate:} \{'node\_name': 'Wall Anchor', 'description': 'Wall Anchor: A hardware product type used to securely fasten objects to hollow walls or masonry surfaces ...'\}

\vspace{0.5em}
\textbf{Choose one action from:}
 \vspace{-1em}
    \begin{itemize} \setlength\itemsep{-0.5em}
        \item ADD: Candidate is not in the KG and not a synonym of an existing node.
        \item MERGE: Candidate already exists in KG or is a synonym of an existing node, and the existing node name is the preferred form.
        \item REPLACE: Candidate is a synonym of an existing node, but the candidate name is the preferred or more standard form, so it should replace the existing node name.
        \item DISCARD: Candidate is invalid, vague, or unreliable.
    \end{itemize}
\vspace{-1em}

\textbf{Preferred Naming Rules:}
\vspace{-1em}
    \begin{itemize} \setlength\itemsep{-0.5em}
        \item Use the most common international or scientific name unless the local name adds significant value.
        \item If candidate is clearer, more standard, or better aligned with naming conventions, choose REPLACE.
        \item If candidate is less standard or too narrow, use MERGE.
        \item Always preserve other name variants in synonyms.
    \end{itemize}
\vspace{-1em}

\textbf{Output format:}
\vspace{-1em}
    \begin{itemize} \setlength\itemsep{-0.5em}
        \item For ADD or DISCARD: output only the action in uppercase.
        \item For MERGE or REPLACE: output the action in uppercase, followed by a space, then the `node\_id` of the matched existing node.
    \end{itemize}
\vspace{-1em}

\textbf{Example outputs:}
\vspace{-1em}
    \begin{itemize} \setlength\itemsep{-0.5em}
        \item MERGE 1749
        \item REPLACE 3943
        \item ADD
        \item DISCARD
    \end{itemize}
\vspace{-1em}

\hrulefill

\vspace{0.5em}
\textbf{Example output:}
MERGE 4587

\end{casepromptbox}
\caption{KGD prompt example (basic version). Given retrieved KG context (top-$k$ relevant nodes) and a candidate node proposal, KGD selects a single constrained action---\texttt{ADD}, \texttt{MERGE}, \texttt{REPLACE}, or \texttt{DISCARD}---to control KG growth and enforce canonicalization. This is a minimum viable version.} 
\label{fig:prompt-kgd-example-basic}
\end{figure*}

\begin{figure*}[t]
\centering
\begin{casepromptbox}{KGD prompt example (Strict Version)}
\small

\textbf{Instruction:}  
You are a product knowledge graph decision agent.

\vspace{0.5em}
\textbf{Goal:}  
Decide whether to ADD a new node, MERGE with an existing node, REPLACE an existing node name, or DISCARD the candidate.

\vspace{0.5em}
\textbf{Relevant nodes from KG:}  
\{pretty\_nodes\}

\vspace{0.5em}
\textbf{Node Type:} \{node\_type\}

\vspace{0.5em}
\textbf{Candidate:} \{pretty\_candidate\}

\vspace{0.5em}
\textbf{Choose one action from:}
\vspace{-1em}
\begin{itemize} \setlength\itemsep{-0.5em}
    \item ADD: Candidate is a distinct concept not present in the KG.
    \item MERGE: Candidate is the SAME concept as an existing node (true synonym), and the existing node name is the preferred form.
    \item REPLACE: Candidate is the SAME concept as an existing node (true synonym), but the candidate name is the preferred/standard form and should replace the existing node name.
    \item DISCARD: Candidate is invalid, too vague, unreliable, or not a real concept for this Node Type.
\end{itemize}
\vspace{-1em}

\textbf{STRICT SYNONYM POLICY (required for MERGE/REPLACE):}
\vspace{-1em}
\begin{itemize} \setlength\itemsep{-0.5em}
    \item Same meaning and scope (no broader/narrower relationship)
    \item Same real-world referent (can be substituted in a sentence without changing meaning)
    \item Same measurement basis/units/standards if applicable (e.g., ratings, units, dimensions)
\end{itemize}
\vspace{-1em}

\textbf{DO NOT MERGE/REPLACE if any of the following is true:}
\vspace{-1em}
\begin{itemize} \setlength\itemsep{-0.5em}
    \item Candidate is related but not identical (association $\neq$ synonymy)
    \item Candidate is a subtype/supertype, variant, feature level, or category of the existing node
    \item Candidate differs by metric vs attribute (e.g., ``size'' vs ``dimensions''; ``weight'' vs ``mass'')
    \item Candidate differs by standard/rating/threshold (e.g., ``splash resistant'' vs a specific IP rating)
    \item Candidate differs by mechanism (e.g., over-voltage vs over-current vs surge vs thermal protection)
    \item Candidate is ambiguous while the existing node is specific, or vice versa
\end{itemize}
\vspace{-1em}

\textbf{DISCARD RULES:}
\vspace{-1em}
\begin{itemize} \setlength\itemsep{-0.5em}
    \item Candidate is not a valid term for this Node Type
    \item Candidate is overly vague or marketing-only with no clear definition
    \item Candidate conflicts with common technical meaning or is unreliable
\end{itemize}
\vspace{-1em}

\textbf{Preferred Naming Rules:}
\vspace{-1em}
\begin{itemize} \setlength\itemsep{-0.5em}
    \item Use the most common international or scientific name unless a local name adds value
    \item If candidate is clearer or more standard, choose REPLACE; otherwise MERGE
    \item Preserve other variants as synonyms (handled outside this decision)
\end{itemize}
\vspace{-1em}

\textbf{Decision procedure:}
\vspace{-1em}
\begin{itemize} \setlength\itemsep{-0.5em}
    \item 1) Validate candidate (otherwise DISCARD)
    \item 2) Check each relevant KG node for strict synonymy
    \item 3) If a strict synonym match exists:
    \begin{itemize} \setlength\itemsep{-0.1em}
        \item If existing name preferred $\rightarrow$ MERGE with that node\_id
        \item If candidate name preferred $\rightarrow$ REPLACE that node\_id
    \end{itemize}
    \item 4) If no strict synonym match exists $\rightarrow$ ADD
\end{itemize}
\vspace{-1em}

\textbf{Output format:}
\vspace{-1em}
\begin{itemize} \setlength\itemsep{-0.5em}
    \item For ADD or DISCARD: output only the action in uppercase
    \item For MERGE or REPLACE: output action + space + node\_id
\end{itemize}
\vspace{-1em}

\textbf{Example outputs:}
\vspace{-1em}
\begin{itemize} \setlength\itemsep{-0.5em}
    \item MERGE 1749
    \item REPLACE 3943
    \item ADD
    \item DISCARD
\end{itemize}
\vspace{-1em}

\hrulefill

\vspace{0.5em}
\textbf{Example output:}  
MERGE 4587

\end{casepromptbox}
\caption{KGD prompt example (strict version). This version has the strictest filtering rules: it lists in detail the situations that cannot be merged (such as different mechanisms, metrics vs. attributes, different standards, etc.).}
\label{fig:prompt-kgd-example-strict}
\end{figure*}

\begin{figure*}[t]
\centering
\begin{casepromptbox}{KGD prompt example (Remove \texttt{Discard} Version)}
\small

\textbf{Instruction:}  
You are a product knowledge graph decision agent.

\vspace{0.5em}
\textbf{Goal:}  
Decide whether to ADD a new node, MERGE with an existing node, or REPLACE an existing node name.

\vspace{0.5em}
\textbf{Relevant nodes from KG:}  

\{pretty\_nodes\}

\vspace{0.5em}
\textbf{Node Type:} \{node\_type\}

\vspace{0.5em}
\textbf{Candidate:} \{pretty\_candidate\}

\vspace{0.5em}
\textbf{Choose one action from:}
\vspace{-1em}
\begin{itemize} \setlength\itemsep{-0.5em}
    \item ADD: Candidate is a distinct concept not present in the KG.
    \item MERGE: Candidate is the SAME concept as an existing node (true synonym), and the existing node name is the preferred form.
    \item REPLACE: Candidate is the SAME concept as an existing node (true synonym), but the candidate name is the preferred/standard form and should replace the existing node name.
\end{itemize}
\vspace{-1em}

\textbf{STRICT SYNONYM POLICY (required for MERGE/REPLACE):}
\vspace{-1em}
\begin{itemize} \setlength\itemsep{-0.5em}
    \item Same meaning and scope (no broader/narrower relationship)
    \item Same real-world referent (can be substituted in a sentence without changing meaning)
    \item Same measurement basis/units/standards if applicable (e.g., ratings, units, dimensions)
\end{itemize}
\vspace{-1em}

\textbf{DO NOT MERGE/REPLACE if any of the following is true (choose ADD instead):}
\vspace{-1em}
\begin{itemize} \setlength\itemsep{-0.5em}
    \item Candidate is related but not identical (association $\neq$ synonymy)
    \item Candidate is a subtype/supertype, variant, feature level, or category of the existing node
    \item Candidate differs by metric vs attribute (e.g., ``size'' vs ``dimensions''; ``weight'' vs ``mass'')
    \item Candidate differs by standard/rating/threshold (e.g., ``splash resistant'' vs a specific IP rating unless explicitly defined as equivalent)
    \item Candidate differs by mechanism (e.g., over-voltage vs over-current vs surge vs thermal protection)
    \item Candidate is ambiguous while the existing node is specific, or vice versa
\end{itemize}
\vspace{-1em}

\textbf{Preferred Naming Rules (apply only after synonymy is confirmed):}
\vspace{-1em}
\begin{itemize} \setlength\itemsep{-0.5em}
    \item Use the most common international or scientific name unless a local name adds significant value.
    \item If candidate is better to be a representative of synonyms, choose REPLACE; otherwise choose MERGE.
    \item Preserve other name variants as synonyms (handled outside this decision).
\end{itemize}
\vspace{-1em}

\textbf{Decision procedure:}
\vspace{-1em}
\begin{enumerate} \setlength\itemsep{-0.5em}
    \item Evaluate candidate validity and relevance to Node Type.
    \item Check each relevant KG node for strict synonymy using the STRICT SYNONYM POLICY.
    \item If a strict synonym match exists:
    \begin{itemize} \setlength\itemsep{-0.2em}
        \item If existing name preferred $\rightarrow$ MERGE with that node\_id
        \item If candidate name preferred $\rightarrow$ REPLACE that node\_id
    \end{itemize}
    \item If no strict synonym match exists $\rightarrow$ ADD
\end{enumerate}
\vspace{-1em}

\textbf{Output format:}
\vspace{-1em}
\begin{itemize} \setlength\itemsep{-0.5em}
    \item For ADD: output only \texttt{ADD}
    \item For MERGE or REPLACE: output the action in uppercase, followed by a space, then the \texttt{node\_id} of the matched existing node
\end{itemize}
\vspace{-1em}

\textbf{Example outputs:}
\vspace{-1em}
\begin{itemize} \setlength\itemsep{-0.5em}
    \item MERGE 1749
    \item REPLACE 3943
    \item ADD
\end{itemize}
\vspace{-1em}

\hrulefill

\vspace{0.5em}
\textbf{Example output:}  
MERGE 4587

\end{casepromptbox}
\caption{KGD prompt example (remove discard version). The \texttt{DISCARD} option has been removed from the action space: it is now assumed that all input candidate terms are valid, with the focus shifting solely to deduplication and name standardization.
Logical Focus: The logic now centers on determining whether a candidate term is more standardized than the existing node name (thereby triggering a \texttt{REPLACE} operation); otherwise, it simply performs a \texttt{MERGE}.}
\label{fig:prompt-kgd-example-no-discard}
\end{figure*}

\begin{figure*}[t]
\centering
\begin{casepromptbox}{Product type suggestion prompt}
\small 
    \textbf{Instruction:} Return the simplest, most general product type that accurately represents the item while ensuring clarity and avoiding ambiguity. Follow these rules:
    
    \vspace{-1em}
    \begin{itemize} \setlength\itemsep{-0.7em}
        \item Remove brand names, model numbers, attributes, and marketing language.
        \item Eliminate redundant or situational descriptors.
        \item Standardize technical terms using widely accepted industry vocabulary.
        \item Abstract to the highest-level accurate category (e.g., "Chair" instead of "Office Chair").
        \item Use singular form, unless the product is commonly referred to in plural (e.g., "Scissors", "Pants", "Shoes").
        \item Return `None` for inputs that are unclear, ambiguous, or not valid products.
        \item If the input already meets all criteria, return it unchanged.
    \end{itemize}
    \vspace{-1em}

    \textbf{Additional Guidance:}

    \vspace{-1em}
    \begin{itemize} \setlength\itemsep{-0.7em}
        \item Contextual Clarity: Ensure the abstracted term maintains enough context to avoid ambiguity. For example, if the original product is clearly related to vehicles, use terms like "Vehicle Part" instead of just "Part."
        \item Specificity Check: If the most general term could refer to multiple unrelated categories (e.g., "Panel" could mean a house panel or a motorcycle panel), provide a more specific term that still fits the criteria (e.g., "Motorcycle Panel").
    \end{itemize}
    \vspace{-1em}
    
    Your goal is to return the most common specific type that is still universally understood and accurately descriptive.
    Provide only the product type in English, no extra text or explanation.

    \vspace{0.5em}
    \textbf{Title:} Wireless Mouse Pen 2.4G Bluetooth Optical Pocket Pen Mouse with Stylus Function for Laptop Tablet Phone Stylus Mouse

    \vspace{0.5em}
    \textbf{Description:} Experience a new level of convenience with our innovative 2.4G Bluetooth Wireless Pen Mouse... [truncated for brevity] ...Package Contents: 1 * Wireless Mouse Pen.

    \vspace{0.5em}
    \textbf{Specifications:} \{``brand'': [``Universal''], ``connectivity'': [``Bluetooth''], ...\}

    \hrulefill

    \vspace{0.5em}
    \textbf{Example output:}
    Pen Mouse

\end{casepromptbox}
\caption{Product type suggestion prompt. The model is instructed to strip non-type information and output a concise, canonical product type that is as general as possible while remaining unambiguous; it must output \texttt{None} when evidence is insufficient.}
\label{fig:prompt-product-type}
\end{figure*}

\begin{figure*}[t]
\centering
\begin{casepromptbox}{Attribute key discovery prompt}
\small
\textbf{Instruction:}
Generate a comprehensive table containing key product attributes.

\vspace{0.5em}
\textbf{The attributes must meet these criteria:}
\vspace{-1em}
\begin{itemize} \setlength\itemsep{-0.7em}
\item Essential for both sellers and buyers
\item Focused on inherent product characteristics (not logistics, packaging, or SEO)
\item Based on industry standards or common e-commerce practices
\item Includes visual and functional attributes that influence purchasing decisions
\end{itemize}
\vspace{-1em}

\textbf{Table format:}
\vspace{-1em}
\begin{itemize} \setlength\itemsep{-0.7em}
\item Attribute Name: Standard Attribute Name
\item Description: What the attribute represents
\item Examples: Examples of Standard Values (comprehensive)
\end{itemize}
\vspace{-1em}

\textbf{Output constraints:}
\vspace{-1em}
\begin{itemize} \setlength\itemsep{-0.7em}
\item Provide only the table, no extra text or explanation.
\item Sort ALL attributes by importance from the buyer's perspective.
\item Start with Brand as the top row, followed by other attributes in importance order.
\item Use proper title case capitalization for all values in Attribute Name and Examples
      (e.g., \texttt{Universal}, \texttt{Infrared}, \texttt{Lithium-Ion}, not \texttt{universal}, \texttt{infrared}, \texttt{lithium-ion}).
\item Provide values in examples at least 5 if applicable.
\item Keep example values less than 10.
\end{itemize}
\vspace{-1em}

\textbf{Product Type:} Pen Mouse

\vspace{0.25em}
\textbf{Product Type Description:}
A hybrid input device that combines the form factor and precision of a stylus pen with the functionality of a computer mouse, designed for enhanced control in graphic design, digital note-taking, and precision navigation on touch-enabled surfaces.

\hrulefill

\vspace{0.5em}
\textbf{Example output:}

\vspace{0.5em}
| Attribute Name | Description | Examples |
\\
|----------------|-------------|----------|
\\
| Brand | The Manufacturer Or Company That Produces The Product | Wacom, Apple, Huion, XP-Pen, Logitech |
\\
| Precision | The Level Of Accuracy And Responsiveness In Tracking Movement And Input | 0.01 Mm, 0.05 Mm, 0.1 Mm, 0.2 Mm, 0.5 Mm |
\\
| Pressure Sensitivity | The Number Of Levels The Device Can Detect To Vary Line Thickness Or Opacity Based On Pen Pressure | 2048, 4096, 8192, 16384, 20480 |
\\ 
(truncated for brevity)

\end{casepromptbox}
\caption{Attribute key discovery prompt. The model generates a buyer-centric,
type-specific attribute schema as a Markdown table (Attribute Name, Description,
Examples) sorted by importance and constrained to standardized naming/value formats.}
\label{fig:prompt-attr-key}
\end{figure*}

\section{Online Deployment: Filter and Badge Demos}
\label{app:demos}

\begin{figure}[t]
\centering
\centering
\includegraphics[width=0.5\textwidth]{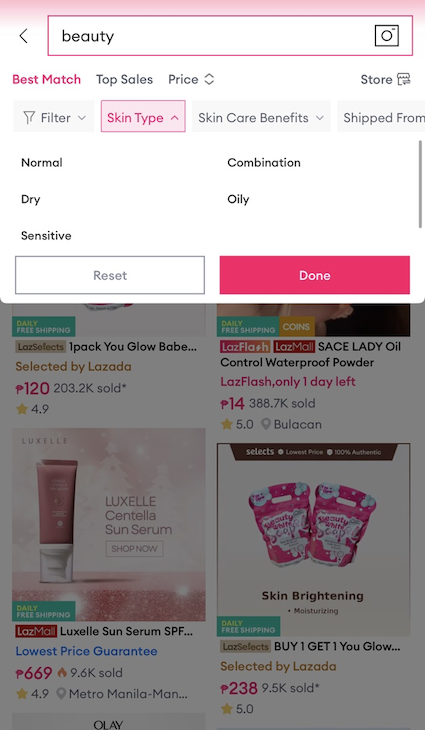}
\caption{Demo of \textit{Filter}: AutoPKG-derived attribute facets surfaced in the UI to refine result sets.}
\label{fig:demo-filter}
\end{figure}

\begin{figure}[t]
\centering
\includegraphics[width=0.5\textwidth]{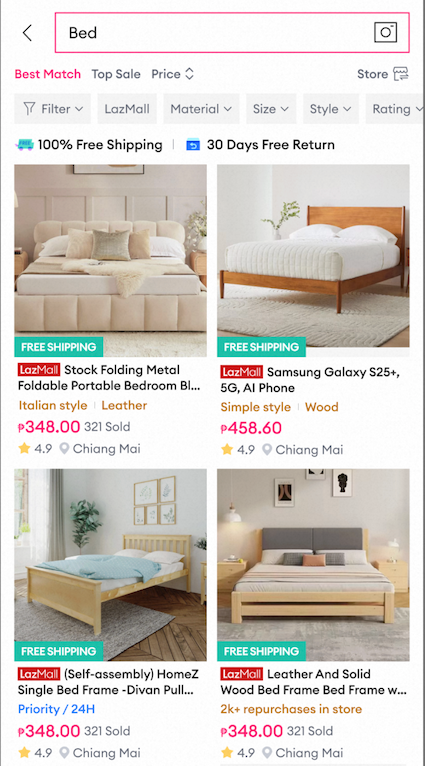}
\caption{Demo of \textit{Badge}: AutoPKG-derived attribute badges shown on product cards.}
\label{fig:demo-badge}
\end{figure}

Figures~\ref{fig:demo-filter} and \ref{fig:demo-badge} illustrate how AutoPKG-derived attributes are surfaced in real user flows. For \textit{Filter}, attribute values are promoted to faceted options (e.g., \textit{Skin Type}) that let users narrow result sets with semantically meaningful constraints. For \textit{Badge}, the same PKG attributes are rendered as concise highlights on product cards (e.g., \textit{Material}, \textit{Style}), improving attribute visibility and reducing the effort needed to compare items.

\end{document}